\documentclass[10pt,twocolumn,letterpaper]{article}

\usepackage[svgnames]{xcolor}
\usepackage{caption}
\usepackage{cvpr}
\usepackage{times}
\usepackage{epsfig}
\usepackage{graphicx}
\usepackage{amsmath}
\usepackage{amssymb}
\usepackage{tikz}

\usepackage{changepage}
\usepackage{algorithm}
\usepackage{algpseudocode}
\usepackage{multirow}
\usepackage{subcaption} 
\pgfdeclarelayer{edgelayer}
\pgfdeclarelayer{nodelayer}
\pgfsetlayers{edgelayer,nodelayer,main}

\tikzstyle{none}=[inner sep=0pt]

\tikzstyle{vertex_default}=[circle,fill=White,draw=Black]
\tikzstyle{edge_forward}=[->,ultra thick]
\tikzstyle{edge_forward_selected}=[->,ultra thick,draw=Red]
\tikzstyle{edge_backward}=[<-,ultra thick,draw=Green]
\tikzstyle{edge_backward_selected}=[<-,ultra thick,draw=Blue]

\usepackage[breaklinks=true,bookmarks=false]{hyperref}

\cvprfinalcopy 

\newcommand\bb[1]{\textbf{#1}}

\begin{document}

\title{Learning Multi-target Tracking with Quadratic Object Interactions}

\author{Shaofei Wang, Charless Fowlkes\\
Dept of Computer Science\\
University of California, Irvine\\
{\tt\small shaofeiw@uci.edu, fowlkes@ics.uci.edu}
}
\maketitle

\begin{abstract}
We describe a model for multi-target tracking based on associating collections
of candidate detections across frames of a video.  In order to model pairwise
interactions between different tracks, such as suppression of overlapping tracks
and contextual cues about co-occurence of different objects, we augment a
standard min-cost flow objective with quadratic terms between detection
variables. We learn the parameters of this model using structured prediction
and a loss function which approximates the multi-target tracking accuracy. We
evaluate two different approaches to finding an optimal set of tracks under 
model objective based on an LP relaxation and a novel greedy extension to
dynamic programming that handles pairwise interactions.  We find the greedy
algorithm achieves equivalent performance to the LP relaxation while being 2-7x
faster than a commercial solver. The resulting model with learned parameters
outperforms existing methods across several categories on the KITTI tracking
benchmark.
\end{abstract}

\section{Introduction}
 \begin{figure}[t]
 \begin{tabular}{cc}
 \includegraphics[clip,trim=22cm 0cm 4cm 3cm,width=0.215\textwidth,height=0.125\textwidth]{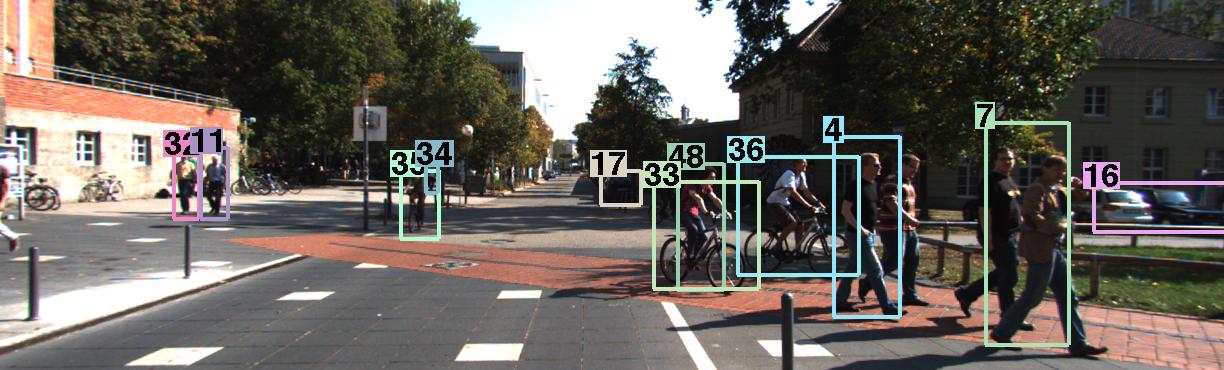}&\includegraphics[clip,trim=22cm 0cm 4cm 3cm,width=0.215\textwidth,height=0.125\textwidth]{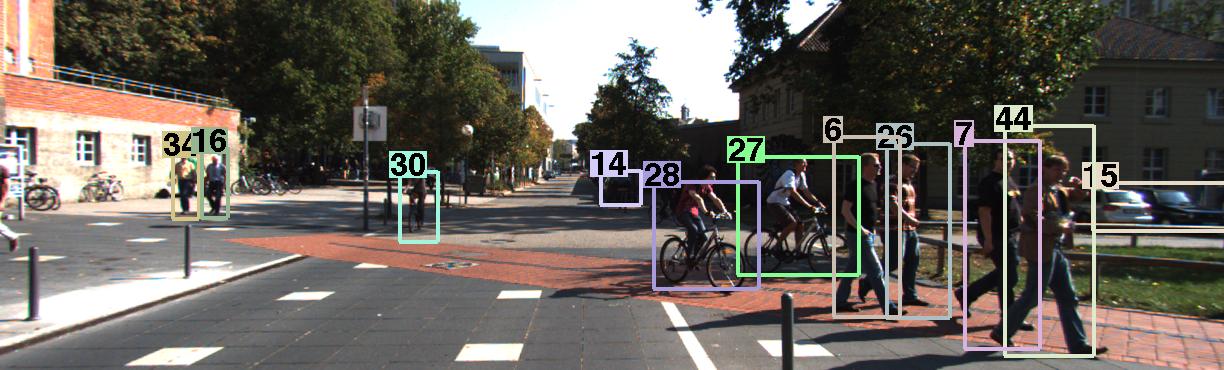}\\
 \includegraphics[clip,trim=0cm 1cm 30cm 5cm,width=0.215\textwidth,height=0.125\textwidth]{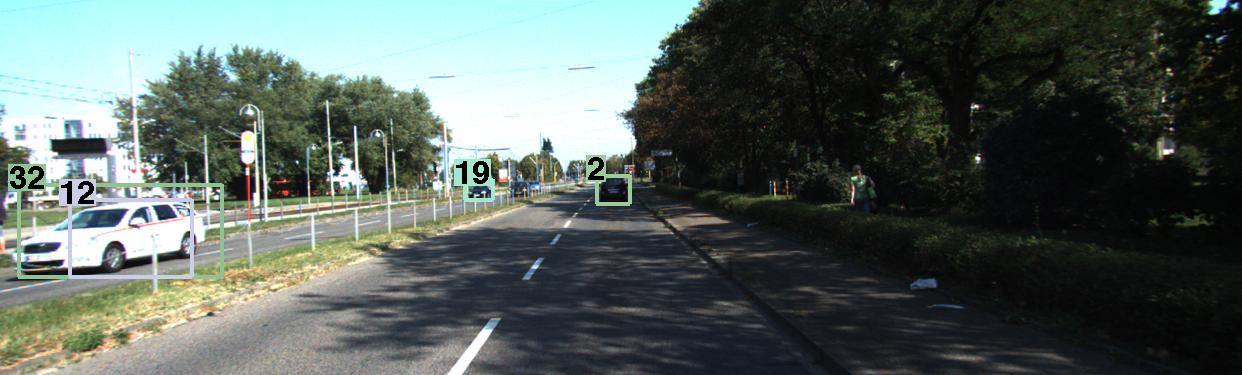}&\includegraphics[clip,trim=0cm 1cm 30cm 5cm,width=0.215\textwidth,height=0.125\textwidth]{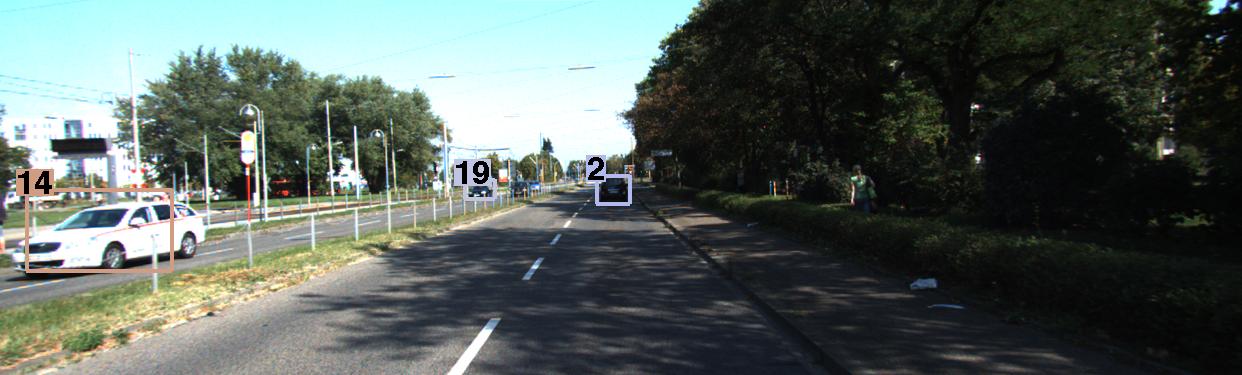}
\end{tabular}
   \caption{We describe a framework for learning parameters of a multi-object
   tracking objective that includes pairwise interactions between objects.
   The left column shows tracking without pairwise interactions.  Our system
   \textit{learns} to enforce both inter-class and intra-class mutual exclusion
   as well as co-occurrence relationship between trajectories.  By
   incorporating pairwise interactions between objects within a frame we are
   able to improve detection performance. }
\end{figure}

Multi-target tracking is a classic topic of research in computer vision.
Thanks to advances of object detector performance in single, still images,
"tracking-by-detection" approaches that build tracks on top of a collection
of candidate object detections have shown great promise.  Tracking-by-detection
avoids some problems such as drift and is often able to recover from extended
periods of occlusion since it is ``self-initializing''.  Finding an optimal set
of detections corresponding to each track is often formulated as a discrete
optimization problem of finding low-cost paths through a graph of candidate
detections for which there are often efficient combinatorial algorithms (such
as min-cost matching or min-cost network-flow) that yield globally optimal
solutions (\eg, ~\cite{10.1109/CVPR.2008.4587584,PirsiavashRF_CVPR_2011}).

Tracking by detection is somewhat different than traditional generative
formulations of multi-target tracking, which draw a distinction between the
problem of estimating a latent continuous trajectory for each object from the
discrete per-frame data-association problem of assigning observations (\eg,
detections) to underlying tracks.  Such methods (\eg,
~\cite{Andriyenko:2012:DCO,Milan:2013:DTE,WuThScBe2012}) allow for explicitly
specifying an intuitive model of trajectory smoothness but face a difficult
joint inference problem over both continuous and discrete variables with little
guarantee of optimality.

In tracking by detection, trajectories are implicitly defined by the selected
group of detections.  For example, the path may skip over some frames entirely
due to occlusions or missing detections.  The transition cost of utilizing a
given edge between detections in successive frames thus could be interpreted as
some approximation of the marginal likelihood associated with integrating over
a set of underlying continuous trajectories associated with the corresponding
pair of detections.  This immediately raises difficulties, both in (1) encoding
strong trajectory models with only pairwise potentials and (2) identifying the
parameters of these potentials from training data.

One line of attack is to first group detections in to candidate tracklets and
then perform scoring and association of these
tracklets~\cite{Yang12anonline,Brendel11multiobjecttracking,Wang_2014_CVPR}.
Tracklets allow for scoring much richer trajectory and appearance models while
maintaining some benefits of purely combinatorial grouping.  Another approach
is to attempt to include higher-order constraints directly in a combinatorial
framework~\cite{Butt_2013_ICCV_Workshops,DBLP:journals/corr/ChariLLS14}.
In either case, there are a large number of parameters associated with these
richer models which necessitates application of machine learning techniques.
This is particularly true for (undirected) combinatorial models based on,
\eg network-flow, where parameters are often set empirically by hand. 

In this work, we introduce an extension to the standard min-cost flow tracking
objective that allows us to model pairwise interactions between tracks.  This
allows us to incorporate useful knowledge such as typical spatial relationships
between detections of different objects and suppression of multiple overlapping
tracks of the same object. This quadratic interaction necessitates the
development of approximate inference methods which we describe in Section
\ref{sec:inference}. In Section \ref{sec:learning} we describe an approach to
joint learning of model parameters in order to maximize tracking performance on
a training data set using techniques for structured
prediction~\cite{Taskar03M3N}.  Structured prediction has been applied in
tracking to learning inter-frame affinity
metrics~\cite{Kim:2012:OMT:2482048.2482058} and
association~\cite{lou_11_structured} as well as a variety of other learning
tasks such as fitting CRF parameters for segmentation~\cite{export:67945} and
word alignment for machine
translation~\cite{Lacoste-Julien:2006:WAV:1220835.1220850}.  To our best
knowledge, the work presented here is unique in utilizing discriminative
structured prediction to \textit{jointly} learn the complete set of parameters
of a tracking model from labeled data, including track birth/death bias, transition
affinities, and multi-object contextual relations.  We conclude with
experimental results (Section \ref{sec:experiments}) which demonstrate that the
learned quadratic model and inference routines yield state of the art
performance on multi-target, multi-category object tracking in urban scenes.

\section{Model}
\label{sec:model}
We begin by formulating multi-target tracking and data association as a
min-cost flow network problem equivalent to that of
~\cite{10.1109/CVPR.2008.4587584}, where individual tracks are described by a
first-order Markov Model whose state space is spatial-temporal locations in
videos.  This framework incorporates a state transition likelihood that
generates transition features in successive frames, and an observation
likelihood that generates appearance features for objects and background.

\subsection{Tracking by Min-cost Flow}
For a given video sequence, we consider a discrete set of candidate object
detection sites $V$ where each candidate site $x=(l,\sigma,t)$ is described 
by its location, scale and frame number.  We write $\Phi = \{\phi_a(x) | x \in
V\}$ for the image evidence (appearance features) extracted at each
corresponding spatial-temporal location in a video.  A single object track
consists of an ordered set of these detection sites: $T = \{x_1, ...  , x_n\}$,
with strictly increasing frame numbers.

We model the whole video by a collection of tracks $\mathcal{T} = \{T_1, ...
, T_k\}$, each of which independently generates foreground object appearances
at the corresponding sites according to distribution $p_{fg}(\phi_a)$ while the
remaining site appearances are generated by a background distribution
$p_{bg}(\phi_a)$.  Each site can only belong to a single track. Our task is to
infer a collection of tracks that maximize the posterior probability
$P(\mathcal{T}|\Phi)$ under the model.  Assuming that tracks behave
independently of each other and follow a first-order Markov model, we can write
an expression for MAP inference:
\begin{align}
\mathcal{T}^* &= \underset{\mathcal{T}}{\operatorname{argmax}} \prod_{T \in \mathcal{T}} P(\Phi|T) P(T) \nonumber \\
\begin{split}
&= \underset{\mathcal{T}}{\operatorname{argmax}} \big( \prod_{T \in \mathcal{T}} \prod_{x \in T} l(\phi_a(x)) \big) \times \\
& \prod_{T \in \mathcal{T}} \big( p_s(x_1) p_e(x_N) \prod_{i=1}^{N-1} p_t(x_{i+1}|x_i) \big)
\end{split}\label{eqn:maptrack}
\end{align}
where
\begin{align}
l(\phi_a(x)) = \frac{p_{fg}(\phi_a(x))}{p_{bg}(\phi_a(x))} \nonumber
\end{align}
is the appearance likelihood ratio that a specific location $x$ corresponds to
the object tracked and $p_s$, $p_e$ and $p_t$ represent the likelihoods for
tracks starting, ending and transitioning between given sites. 

The set of optimal tracks can be found by taking the log of
\ref{eqn:maptrack} to yield an integer linear program (ILP) over flow variables $\mathbf{f}$.
\begin{align}
\underset{\mathbf{f}}{\operatorname{min}} &\ \sum_{i} c_i^s f_i^s + \sum_{ij \in E} c_{ij} f_{ij} + \sum_{i} c_i f_i + \sum_{i} c_i^t f_i^t 
\label{eqn:mincostflow} \\
\text{s.t.}&\quad f_i^s + \sum_j f_{ji} = f_i = f_i^t + \sum_j f_{ij} \nonumber\\
& f_i^s, f_i^t, f_i, f_{ij} \in \{0,1\} \nonumber 
\end{align}
where $E$ is the set of valid transitions between sites in successive 
frames and the costs are given by
\begin{align}
\begin{split}
c_i = -\log l(\phi_a(x)), c_{ij} = -\log p(x_j | x_i)\\
c_i^s = -\log p_s(x_i), c_i^t = -\log p_t(x_i) 
\label{eqn:potentials}
\end{split}
\end{align}
This ILP is a well studied problem known as minimum-cost network flow~\cite{SSP}.
The constraints satisfy the \textit{total unimodularity} property and thus can
be solved exactly using any LP solver or via various efficient specialized solvers, 
including network simplex, successive shortest path and push-relabel with
bisectional search~\cite{10.1109/CVPR.2008.4587584}.

While these approaches yield globally optimal solutions, the authors
of~\cite{PirsiavashRF_CVPR_2011} consider even faster approximations based on
multiple rounds of dynamic programming (DP).  In particular, the successive shortest
paths algorithm (SSP) finds optimal flows by applying Dijkstra's algorithm on a
residual graph constructed from the original network in which some edges corresponding
to instanced tracks have been reversed. This can be implemented by performing
multiple forward and backward passes of dynamic programming (see Appendix for details).  $\cite{PirsiavashRF_CVPR_2011}$ found that two or even
one pass of DP often performs nearly as well as SSP in practical tracking 
scenarios.  In our experiments we evaluate several of these variants.

\subsubsection{Track interdependence}
The aforementioned model assumes tracks are independent of each other, which is
not always true in practice. A key contribution of our work is showing that
pairwise relations between tracks can be integrated into the model to improve
tracking performance.  In order to allow interactions between multiple objects,
we add a pairwise cost term denoted $q_{ij}$ and $q_{ji}$ for jointly
activating a pair of flows $f_i$ and $f_j$ corresponding to detections at 
sites $x_i = (l_i,\sigma_i,t_i)$ and $x_j=(l_j,\sigma_j,t_j)$.  An intuitive
example of $q_{ij}$ and $q_{ji}$ would be penalty for overlap locations or a
boost for co-occurring objects.  We only consider pairwise interactions between
pairs of sites in the same video frame which we denote by $EC = \{ij : t_i = t_j\}$.
Adding this term to \ref{eqn:mincostflow} yields an Integer Quadratic Program
(IQP):
\begin{align}
\begin{split}
\underset{\mathbf{f}}{\operatorname{min}}& \sum_{i} c_i^s f_i^s + \sum_{ij \in E} c_{ij} f_{ij} + \sum_{i} c_i f_i \\
&+ \sum_{ij \in EC } q_{ij} f_i f_j + \sum_{i} c_i^t f_i^t
\end{split} \label{eqn:quadflow} \\
\text{s.t.}&\quad f_i^s + \sum_j f_{ji} = f_i = f_i^t + \sum_j f_{ij} \nonumber\\
& f_i^s, f_i^t, f_i, f_{ij} \in \{0,1\} \nonumber 
\end{align}
The addition of quadratic terms makes this objective hard to solve in general.
In the next section we discuss two different approximations for finding high
quality solutions $\mathbf{f}$.  In Section~\ref{sec:learning} we describe how
the costs $\mathbf{c}$ can be learned from data.

\section{Inference}
\label{sec:inference}
Now we describe different methods to conduct tracking inference (finding the
optimal flows $\mathbf{f}$).  These inference routines are used both for
predicting a set of tracks at test time as well as optimizing parameters during
learning (see Section~\ref{sec:learning}).

As mentioned in previous section, for traditional min-cost network flow problem defined in Equation~\ref{eqn:mincostflow} there exists various efficient solvers that explores its \textit{total unimodularity} property to find the global optimum. We employ MOSEK's built-in network simplex solver in our experiments,  as other alternative algorithms yield exactly the same solution.

In contrast, finding the global minimum of the IQP problem~\ref{eqn:quadflow}
is NP-hard~\cite{ANasser14} due to the quadratic terms.  We evaluate two
different schemes for finding high-quality approximate solutions.  The first is
a standard approach of introducing auxiliary variables and relaxing the
integral constraints to yield a linear program (LP) that lower-bounds the 
original objective.  We also consider a greedy approximation based on successive
rounds of dynamic programming that also yields good solutions while avoiding
the expense of solving a large scale LP.

\subsection{LP Relaxation and Rounding}
If we relax the integer constraints and deform the costs as necessary to make
the objective convex, then the global optimum of~\ref{eqn:quadflow} can be
found in polynomial time. For example, one could apply Frank-Wolfe algorithm
to optimize the relaxed, convexified QP while simultaneously keeping track of
good integer solutions~\cite{TangECCV14}.  However, for real-world tracking
over long videos, the relaxed QP is still quite expensive. Instead
we follow the approach proposed by Chari
\etal~\cite{DBLP:journals/corr/ChariLLS14}, reformulating the IQP as an
equivalent ILP problem by replacing the quadratic terms $f_i f_j$ with a set of
auxiliary variables $u_{ij}$:
\begin{align}
\begin{split}
\underset{\mathbf{f}}{\operatorname{min}}
\sum_{i} c_i^s f_i^s + \sum_{ij \in E} c_{ij} f_{ij} + \sum_{i} c_i f_i \\
+ \sum_{ij \in EC } q_{ij} u_{ij} + \sum_{i} c_i^t f_i^t
\end{split} \label{eqn:lprelax} \\
\begin{split}
\text{s.t.} \; f_i^s, f_i^t, f_i, f_j, f_{ij}, u_{ij} \in \{0,1\} \\
f_i^s + \sum_j f_{ji} = f_i = f_i^t + \sum_j f_{ij}
\end{split} \nonumber \\
\begin{split}
u_{ij} \le f_i, u_{ij} \le f_j \nonumber \\
f_i + f_j \le u_{ij} + 1 \nonumber
\end{split}
\end{align}
The new constraint sets enforce $u_{ij}$ to be $1$ only when $f_i$ and $f_j$
are both $1$. By relaxing the integer constraints, program
\ref{eqn:lprelax} can be solved efficiently via large scale LP solvers
such as CPLEX or MOSEK. 

During test time we would like to predict a discrete set of tracks. This
requires rounding the solution of the relaxed LP to some solution that
satisfies not only integer constraints but also flow constraints.
~\cite{DBLP:journals/corr/ChariLLS14} proposed two rounding heuristics: a
Euclidean rounding scheme that minimizes $\| \mathbf{f} - \widehat{\mathbf{f}}
\|^2$ where $\widehat{\mathbf{f}}$ is the non-integral solution given by the LP
relaxation.  When $\mathbf{f}$ is constrained to be binary, this objective
simplifies to a linear function $(\mathbf{1}-2 \widehat{\mathbf{f}})^T \mathbf{f} + \|
\widehat{\mathbf{f}} \|^2$, which can be optimized using a standard linear
min-cost flow solver.  Alternately, one can use a linear under-estimator of
\ref{eqn:quadflow} similar to the Frank-Wolfe algorithm:
 \begin{align}
\begin{split}
&\sum_{i} c_i^s f_i^s + \sum_{ij \in E} c_{ij} f_{ij} +\\
&\sum_{i} ( c_i + \sum_{ij \in EC} q_{ij} \widehat{u}_{ij} + \sum_{ji \in EC} q_{ji} \widehat{u}_{ji} ) f_i + \sum_{i} c_i^t f_i^t
\end{split}
\end{align}
Both of these rounding heuristics are linear functions 
subject to the original integer and flow constraints and thus can be solved as
an ordinary min-cost network flow problem. In our experiments we execute both
rounding heuristics and choose the solution with lower cost.

\subsection{Greedy Sequential Search}
We now describe a simple greedy algorithm inspired by the combination of
dynamic programming and non-maximal suppression proposed in
~\cite{PirsiavashRF_CVPR_2011}.  We carry out a series of rounds of dynamic
programming to find the shortest path between source and sink nodes. In each
round, once we have identified a track, we update the (unary) costs associated
with all detections to include the effect of the pairwise quadratic interaction
term of the newly activated track (\eg suppressing overlapping detections,
boosting the scores of commonly co-occurring objects).  This is analogous to 
greedy algorithms for maximum-weight independent set where the elements are
paths through the network. 

\begin{algorithm}{}
\begin{minipage}{0.45\textwidth}
\caption{DP with pairwise Cost Update}
\label{alg:alg1}
\begin{algorithmic}[1]
\State \textbf{Input}: A Directed-Acyclic-Graph $G$ with edge weights $c_i,c_{ij}$
\State initialize $\mathcal{T} \leftarrow \emptyset$
\Repeat
  \State Find shortest start-to-end path $p$ on $G$ 
  \State $track\_cost = cost(p)$
  \If {$track\_cost < 0$}
    \ForAll{locations $x_i$ in $p$}
      \State $c_j = c_j + q_{ij} + q_{ji}$ for all $ij,ji \in EC$
      \State $c_i = +\infty$
    \EndFor
    \State $\mathcal{T} \leftarrow \mathcal{T} \cup p$
  \EndIf
\Until{$track\_cost \ge 0$}
\State \textbf{Output}: track collection $\mathcal{T}$
\end{algorithmic}
\end{minipage}
\end{algorithm}

In the absence of quadratic terms, this algorithm corresponds to the 1-pass DP
approximation of the successive-shortest paths (SSP) algorithm.  Hence it does
not guarantee an optimal solution, but, as we show in the experiments, it
performs well in practice.  A practical implementation difference (from the
linear objective) is that updating the costs with the quadratic terms when a
track is instanced has the unfortunate effect of invalidating cost-to-go
estimates which could otherwise be cached and re-used between successive rounds
to accelerate the DP computation.

Interestingly, the greedy approach to updating the pairwise terms can also be
used with a 2-pass DP approximation to SSP where backward passes subtract 
quadratic penalties.  We describe the details of our implementation of the
2-pass algorithm in the Appendix.  We found the 1-pass approach superior as
the complexity and runtime grows substantially for multi-pass DP with pairwise
updates.

\section{Tracking Features and Potentials}

In order to learn the tracking potentials ($\mathbf{c}$ and $\mathbf{q}$) we
parameterize the flow cost objective by a vector of weights $\mathbf{w}$ and a set of
features $\Psi(X,\mathbf{f})$ that depend on features extracted from the video,
the spatio-temporal relations between candidate detections, and which tracks
are instanced.  With this linear parameterization we write the cost of a given
flow as $C(\mathbf{f}) = -\mathbf{w}^T \Psi(X,\mathbf{f})$ where the negative sign is a
useful convention to convert the minimization problem into a maximization.
The vector components of the weight and feature vector are given by:
\begin{align}
\mathbf{w} =
\begin{bmatrix}
w_S\\ w_t\\ w_s\\ w_a\\ w_E
\end{bmatrix}
\indent \Psi(X,\mathbf{f}) = 
\begin{bmatrix}
\sum_i \phi_S(x_i^s) f_i^s \\ \sum_{ij \in E} \psi_t(x_i, x_j) f_{ij} \\ \sum_{ij \in EC} \psi_s(x_i, x_j) f_i f_j \\ \sum_i \phi_a(x_i) f_i \\ \sum_i \phi_E(x_i^t) f_i^t
\end{bmatrix}
\end{align}
Here $w_a$ represents local appearance template for the tracked objects of
interest, $w_t$ represents weights for transition features, $w_s$ represents
weights for pairwise interactions, $w_S$ and $w_E$ represents weights
associated with track births and deaths. $\phi_a(x_i)$ is
the image feature at spatial-temporal location $x_i$, $\psi_t(x_i, x_j)$
represents the feature of transition from location $x_i$ to location $x_j$,
$\psi_s(x_i, x_j)$ represents the feature of pairwise interaction between location
$x_i$ and $x_j$ that are in the same frame, $\phi_S(x_i^s)$ represents
feature of birth node to location $x_i$ and $\phi_E(x_i^t)$ represents feature
of location $x_i$ to sink node.

\textbf{Local appearance model:} We make use of an off-the-shelf detector
to capture local appearance.  Our local appearance feature thus consists 
of the detector score along with a constant 1 to allow for a variable 
bias.

\textbf{Transition model:} We use a simple motion model (described in
Section \ref{sec:experiments}) to predict candidate windows' locations in future
frames; we connect a candidate $x_i$ at time $t_i$ with another candidate $x_j$
at a later time $t_i+n$, only if the overlap ratio between $x_i$'s predicted
window at $t_i+n$ and $x_j$'s window at $t_i+n$ exceeds $0.3$. The overlap
ratio is defined as two windows' intersection over their union. We use this
overlap ratio as a feature associated with each transition link. The transition
link's feature will be 1 if this ratio is lower than 0.5, and 0 otherwise. In
our experiments we allow up to $7$ frames occlusion for all the network-flow
methods. We append a constant 1 to this feature and bin these features according 
to the length of transition. This yields a $16$ dimensional feature for each transition 
link.

\textbf{Birth/death model:} 
In applications with static cameras it can be useful to learn a spatially
varying bias to model where tracks are likely to appear or disappear.
However, videos in our experiments are all captured from a moving vehicle,
 we thus use a single constant value 1 for the birth and death features.

\textbf{Pairwise interactions:}
$w_s$ is a weight vector that encodes valid geometric configurations of two
objects. $\psi(x_i, x_j)$ is a discretized spatial-context feature
that bins relative location of detection window at location $x_i$ and window at
location $x_j$ into one of the $D$ relations including on top of, above, below,
next-to, near, far and overlap (similar to the spatial context
of~\cite{DesaiRF_ICCV_2009}). To mimic the temporal NMS described in
\cite{PirsiavashRF_CVPR_2011} we add one additional relation, strictly overlap,
which is defined as the intersection of two boxes over the area of the first box;
we set the corresponding feature to 1 if this ratio is greater than 0.9 and 0 otherwise. 
Now assume that we have $K$ classes of objects in
the video, then $w_s$ is a $DK^2$ vector, \ie $w_s = [w_{s11}^T, w_{s12}^T,
..., w_{sij}^T, ... , w_{sKK}^T]^T$, in which $w_{sij}$ is a length of $D$
column vector that encodes valid geometric configurations of object of class
$i$ w.r.t. object of class $j$. In such way we can capture intra- and
inter-class contextual relationships between tracks.

\section{Learning}
\label{sec:learning}

We formulate parameter learning of tracking models as a structured prediction
problem.  With some abuse of notation, assume we have $N$ training videos
$(X_n, \mathbf{f}_n) \in \mathcal{X} \times \mathcal{F}, n = 1,..., N$. Given
ground-truth tracks in training videos specified by flow variables $\mathbf{f}_n$,
we discriminatively learn tracking model parameters $w$ using a structured SVM with
margin rescaling:
\begin{align}
&\mathbf{w}^* = \underset{\mathbf{w},\xi_n \ge 0}{\operatorname{argmin}} \ \frac{1}{2} \| \mathbf{w} \| ^2  + C \sum_n \xi_n \label{eqn:structsvm}\\
\text{s.t.} &\ \forall n, \widehat{\mathbf{f}}, \langle \mathbf{w}, \bigtriangleup \Psi(X_n,\mathbf{f}_n, \widehat{\mathbf{f}})  \rangle \ge L(\mathbf{f}_n, \widehat{\mathbf{f}}) - \xi_n \nonumber
\end{align}
where
\begin{align}
&\bigtriangleup \Psi(X_n,\mathbf{f}_n, \widehat{\mathbf{f}}) = \Psi(X_n,\mathbf{f}_n) - \Psi(X_n,\widehat{\mathbf{f}}) \nonumber
\end{align}
where $\Psi(X_n,\mathbf{f}_n)$ are the features extracted from $n$th training video.
$L(\mathbf{f}_n, \widehat{\mathbf{f}})$ is a loss
function that penalize any difference between the inferred label
$\widehat{\mathbf{f}}$ and the ground truth label $\mathbf{f}_n$. The
constraint on the slack variables $\xi_n$ ensure that we pay a cost for
any training videos in which the flow cost of the ground-truth tracks under
model $w$ is higher than some other incorrect labeling.

\subsection{Cutting plane optimization}
We optimize the structured SVM objective in~\ref{eqn:structsvm} using a
standard cutting-plane method~\cite{Joachims/etal/09a} in which the exponential
number of constraints (one for each possible flow $\widehat{\mathbf{f}}$)
are approximated by a much smaller number of terms.  Given a current 
estimate of $\mathbf{w}$ we find a ``most violated constraint'' for each training
video:
\[
\widehat{\mathbf{f}}_n^* = \underset{\widehat{\mathbf{f}}}{\operatorname{argmax}} \  L(\mathbf{f}_n, \widehat{\mathbf{f}}) - \langle \mathbf{w},  \bigtriangleup \Psi(X_n,\mathbf{f}_n, \widehat{\mathbf{f}}) \rangle
\]
We can then add these constraints to the optimization problem and solve for an
updated $\mathbf{w}$.  This procedure is iterated until no additional constraints are
added to the problem.  In our implementation, at each iteration we add a single
linear constraint which is a sum of violating constraints derived from
individual videos in the dataset which is also a valid cutting
plane constraint~\cite{DesaiRF_ICCV_2009}.

The key subroutine is finding the most-violated constraint for a given video
which requires solving the loss-augmented inference problem (we drop the $n$
subscript notation from here on)
\begin{align}
\widehat{\mathbf{f}}^* &= \underset{\widehat{\mathbf{f}}}{\operatorname{argmin}} \ \langle w,  \Psi(X,\widehat{\mathbf{f}}) \rangle - L(\mathbf{f}, \widehat{\mathbf{f}}) 
\end{align}
As long as the loss function $L(\mathbf{f}, \widehat{\mathbf{f}})$ 
decomposes as a sum over flow variables then this problem has the same form
as our test time tracking inference problem, the only difference being that the
cost of variables in $\mathbf{f}$ is augmented by their corresponding negative
loss.

We note that our two inference algorithms behave somewhat differently when
producing constraints.  The greedy algorithm has no guarantee of finding the
optimal flow for a given tracking problem and hence may not generate all the
necessary constraints for learning $\mathbf{w}$. In contrast, for the LP relaxation, we
have the option of adding constraints corresponding to fractional solutions
(rather than rounding them to discrete tracks). If we use a loss function that
penalizes incorrect non-integral solutions, this may push the structured SVM to
learn parameters that tend to result in tight relaxations.  These scenarios are
termed ``undergenerating'' and ``overgenerating'' respectively by
\cite{Finley/Joachims/08a} since approximate inference is performed over a
subset or superset of the exact space of flows.

\subsection{Loss function}
Now we describe loss functions for multi-target tracking problem. We use a
weighted hamming loss to measure loss between ground truth labels $\mathbf{f}$
and inferred labels $\widehat{\mathbf{f}}$:
\begin{align}
L(\widehat{\mathbf{f}} , \mathbf{f}) = \sum_{f_i \in \mathbf{f}} loss_i \left| f_i -  \widehat{f_i} \right|
\end{align}
where $\{ loss_1, ..., loss_i,..., loss_{|\mathbf{f}|} \}$ is a
vector indicating the penalty for differences between the estimated
flow $\widehat{\mathbf{f}}$ and the ground-truth $\mathbf{f}$.
For example, when $\mathbf{loss} = \mathbf{1}$ it becomes the hamming loss.

\textbf{Transition Loss:} A critical aspect for successful learning is to
define a good $\mathbf{loss}$ vector that closely reassembles major tracking
performance criteria, such as Multiple Object Tracking Accuracy
(MOTA~\cite{Bernardin:2008:EMO:1384968.1453688}). Metrics
such as false positive, false negative, true positive, true negative and
true/false birth/death can be easily incorporated by setting their
corresponding values in $\mathbf{loss}$ to 1. 

By definition, id switches and fragmentations~\cite{Li09learningto} are determined by looking at
labels of two consecutive transition links simultaneously, under such
definition the loss cannot be optimized by our inference routine which only
considers pairwise relations between detections within a frame.  Instead, we
propose a decomposable loss for transition links that attempts to capture
important aspects of MOTA by taking into account the length and localization of
transition links rather than just using a constant (Hamming) loss on mislabeled
links.  We found empirically that careful specification of the loss function
is crucial for learning a good tracking model.

In order to describe our transition loss, let us first denote four types of
transition links: $NN$ is the link from a false detection to another false
detection, $PN$ is the link from a true detection to a false detection, $NP$ is
the link from a false detection to a true detection, $PP^+$ is the link from a
true detection to another true detection with the same identity, and $PP^-$ is
the link from a true detection to another true detection with a different
identity. For all the transition links, we interpolate detections between its
start detection and end detection (if their frame numbers differ more than 1);
the interpolated virtual detections are considered either true virtual
detection or false virtual detection, depending on whether they overlap with a
ground truth label or not.  Loss for different types of transition is defined
as:

\begin{adjustwidth}{1em}{0pt}
1. For $NN$ links, the loss will be (number of true virtual detections + number of false virtual detections)\\
2. For $PN$ and $NP$ links, the loss will be (number of true virtual detections + number of false virtual detections + 1)\\
3. For $PP^+$ links, the loss will be (number of true virtual detections)\\
4. For $PP^-$ links, the loss will be (number of true virtual detections + number of false virtual detections + 2)
\end{adjustwidth}

{\bf Ground-truth flows:} In practice, available training datasets specify
ground-truth bounding boxes that need to be mapped onto ground-truth flow
variables $\mathbf{f_n}$ for each video.  To do this mapping, we first consider
each frame separately, taking the highest scoring detection window that
overlaps a ground truth label as true detection, each true detection will be
assigned a track identity label same as the ground truth label it overlaps.
Next, for each track identity, we run a simplified version of the dynamic
programming algorithm to find the path that claims the largest number of true
detections.  After we iterate through all id labels, any instanced graph 
edge will be a true detection/transition/birth/death while the remainder
will be false.

\section{Experimental results}
\label{sec:experiments}
 \begin{figure}[t]
 \begin{tabular}{cc}
 \includegraphics[clip,trim=5cm 3cm 27cm 5cm,width=0.215\textwidth,height=0.125\textwidth]{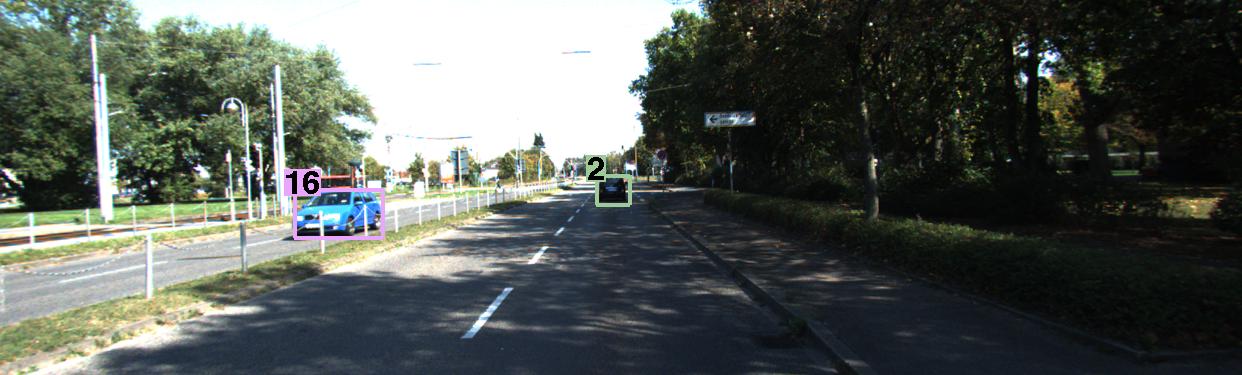}&\includegraphics[clip,trim=5cm 3cm 27cm 5cm,width=0.215\textwidth,height=0.125\textwidth]{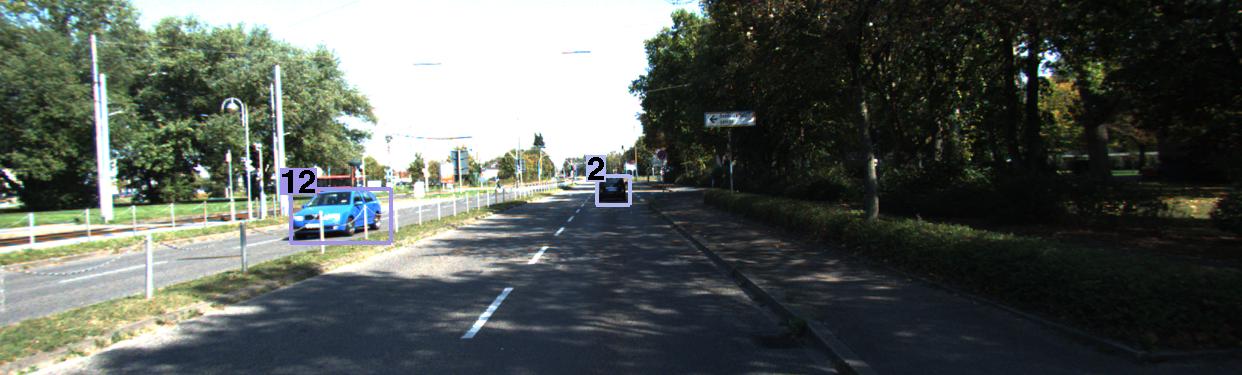}\\
\includegraphics[clip,trim=5cm 3cm 27cm 5cm,width=0.215\textwidth,height=0.125\textwidth]{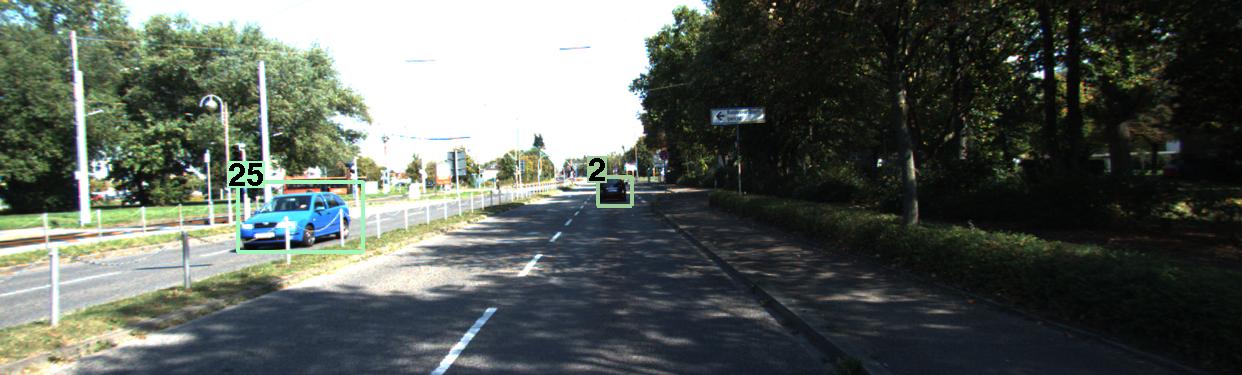} &\includegraphics[clip,trim=5cm 3cm 27cm 5cm,width=0.215\textwidth,height=0.125\textwidth]{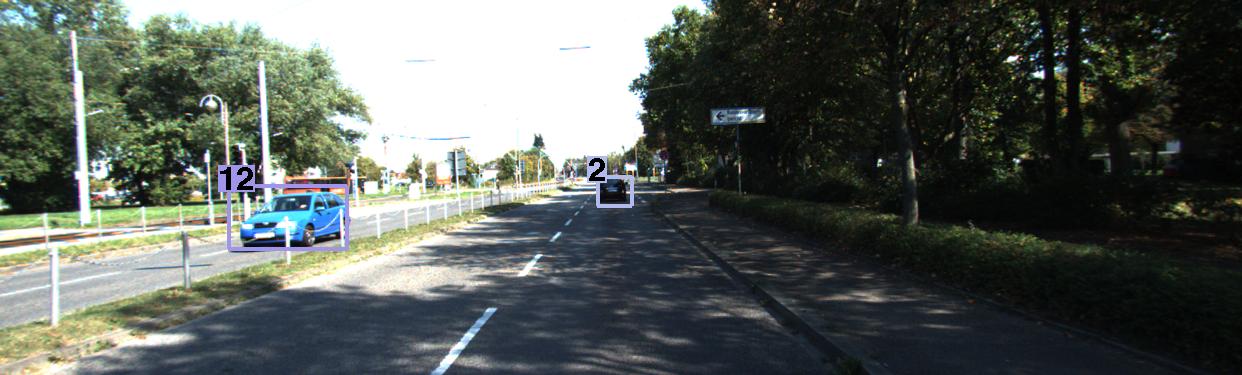}
\end{tabular}
   \caption{Example benefit of soft transition penalty. Left column is an ID
   switch error (IDSW) of the baseline due to removing aggressive transition
   links based on an empirical hard overlap threshold. At right column,
   our model prevents this error by learning a soft penalty function that
   allows for some aggressive transitions to occur.}
\end{figure}
 \begin{figure}[t]
 \begin{tabular}{cc}
 \includegraphics[clip,trim=16cm 5cm 22cm 5cm,width=0.215\textwidth,height=0.125\textwidth]{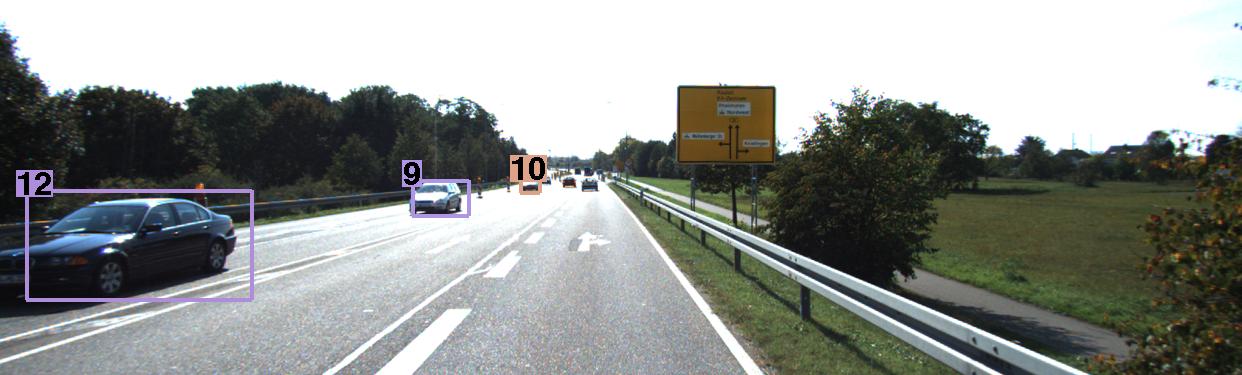}&\includegraphics[clip,trim=16cm 5cm 22cm 5cm,width=0.215\textwidth,height=0.125\textwidth]{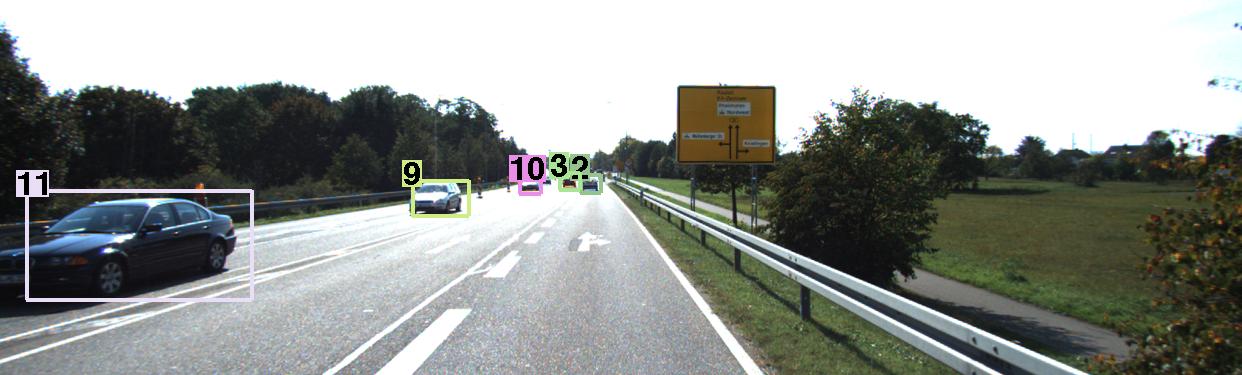}\\
 \includegraphics[clip,trim=16cm 5cm 22cm 5cm,width=0.215\textwidth,height=0.125\textwidth]{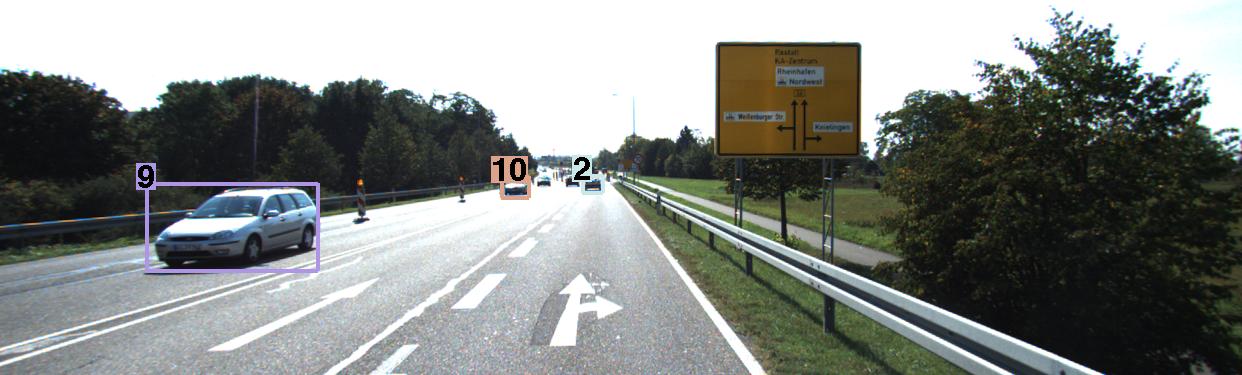}&\includegraphics[clip,trim=16cm 5cm 22cm 5cm,width=0.215\textwidth,height=0.125\textwidth]{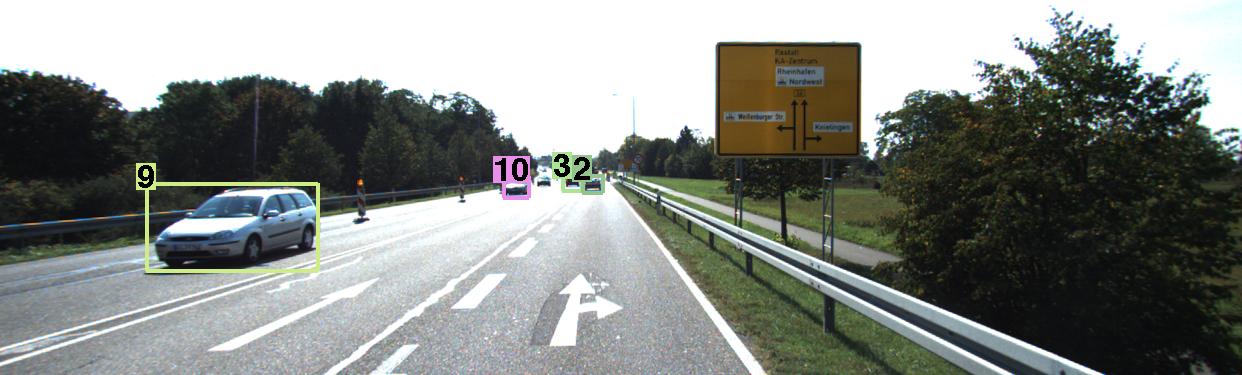}\\
  \includegraphics[clip,trim=16cm 5cm 22cm 5cm,width=0.215\textwidth,height=0.125\textwidth]{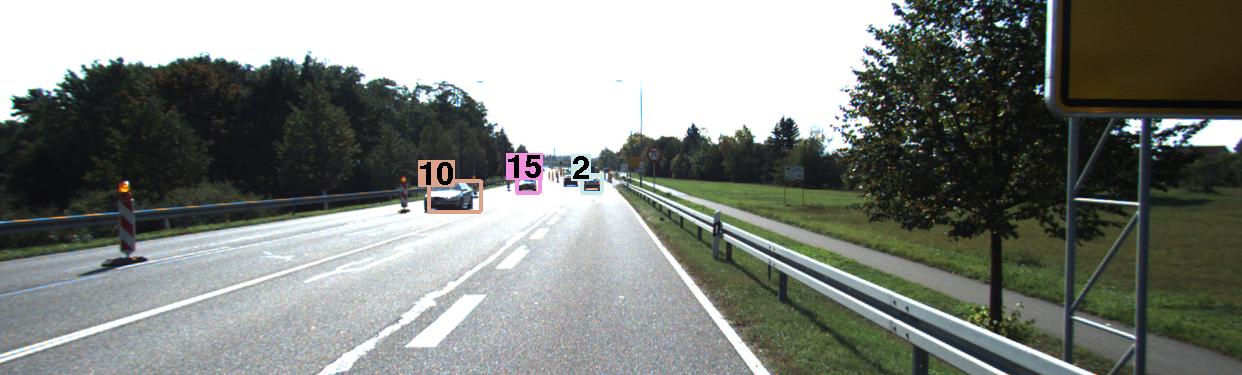}&\includegraphics[clip,trim=16cm 5cm 22cm 5cm,width=0.215\textwidth,height=0.125\textwidth]{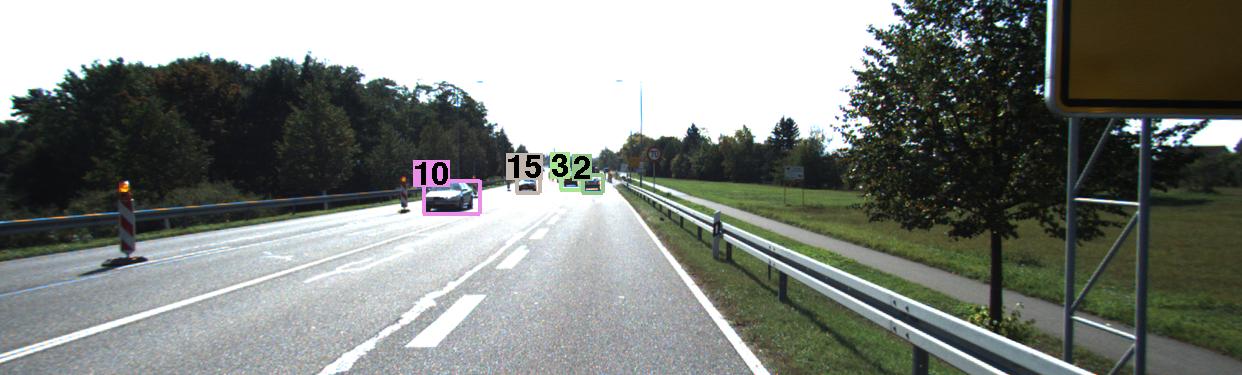}
\end{tabular}
   \caption{Example of track co-occurrence. The right column is the model
   learned with pairwise terms (LP+Flow+Struct), while the left column is
   learned without pairwise terms (SSP+FLow+Struct). Co-occurrence term forces
   both track 2 and 3 to initialize even when the detector responses are weak.}
\end{figure}

\begin{figure}[t]
 \begin{center}
 \begin{tabular}{cc}
 \includegraphics[clip,trim=0cm 0.55cm 0cm 0cm,width=0.45\linewidth]{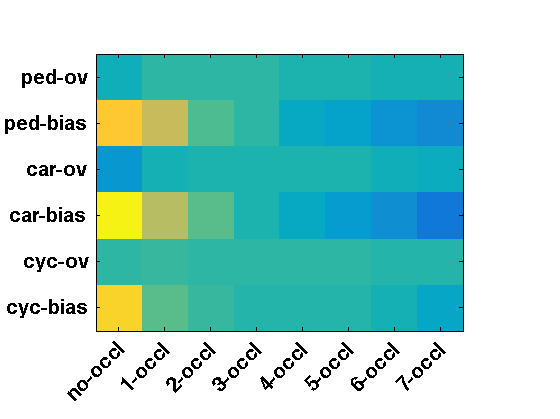} &
 \includegraphics[clip,trim=0cm 1.75cm 1cm 0cm,width=0.45\linewidth]{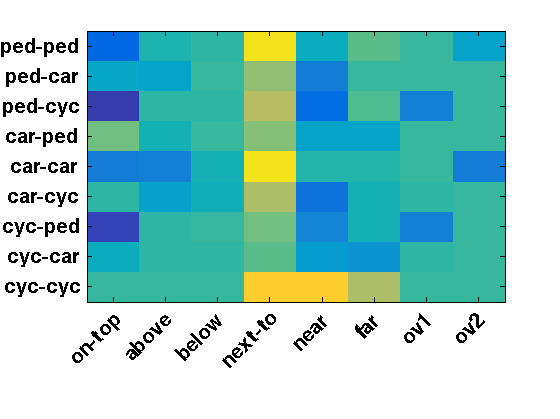} \\
 (a) inter-frame weights & (b) intra-frame weights \\
 \end{tabular}
\end{center}
   \caption{Visualization of the weight vector learned by our method. Yellow
   has small cost, blue has large cost.
   (a) shows transition weights for different length of frame jumps. The
   model encourages transitions to nearby neighboring frames, and penalizes
   long or weak transition links (\ie overlap ratio lower than 0.5).  (b)
   shows learned pairwise contextual weights between objects.  The model
   encourages intra-class co-occurrence when objects are close but penalizes
   overlap and objects on top of others.  Note the strong negative interaction
   learned between cyclist and pedestrian (two classes which are easily
   confused by their respective detectors.).  By exploring contextual cues we
   can make correct prediction on this otherwise confusing configuration.
   }
\end{figure}

\textbf{Dataset:} We have focused our experiments on training sequences of KITTI tracking
benchmark~\cite{Geiger2012CVPR}. KITTI tracking benchmark consists of to 21
training sequences with a total of 8008 frames and 8 classes of labeled
objects; of all the labeled objects we evaluated three categories which had
sufficient number of instances for comparative evaluation: cars, pedestrians
and cyclists.  We use publicly available LSVM~\cite{voc-release4} reference detections and evaluation script\footnote{\url{http://www.cvlibs.net/datasets/kitti/eval_tracking.php}}.
 The evaluation script only evaluates objects that are not too far away and not truncated by more than 15 percent, it also does not consider vans as false positive for cars or sitting
persons as false positive for pedestrians.  The final dataset contains
636 labeled car trajectories, 201 labeled pedestrian trajectories and 37
labeled cyclists trajectories.

\textbf{Training with ambiguous labels:} One difficulty of training on the
KITTI tracking benchmark is that it has special evaluation rules for ground
truth labels such as small/truncated objects and vans for cars, sitting persons
for pedestrians. This is resolved by removing all detection candidates that
correspond to any of these ``ambiguous" ground truth labels during training; in
this way we avoid mining hard negatives from those labels. Also, to speed up
training, we partition full-sized training sequences in to 10-frame-long
subsequences with a 5-frame overlap, and define losses on each subsequence
separately.
\begin{figure}[t]
\begin{center}
   \includegraphics[clip,trim=0cm 6cm 0cm 6cm,width=1.0\linewidth]{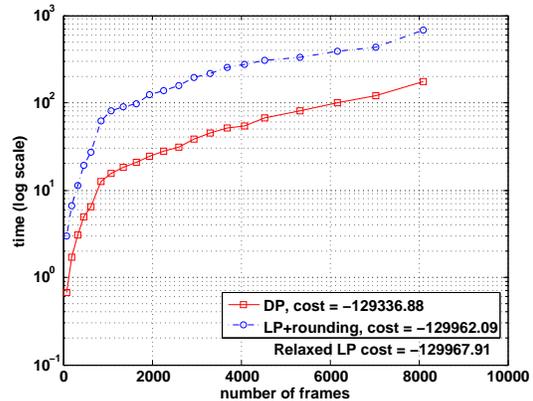}
\end{center}
   \caption{Speed and quality comparison of proposed undergenerating and
   overgenerating approximation. Over the 21 training sequences in KITTI
   dataset, LP+rounding produces cost that is very close to relaxed global
   optimum. DP gives a lower bound that is within 1\% of relaxed global
   optimum, while being 2 to 7 times faster than a commercial LP solver (MOSEK)
   }
\label{fig:fig5}
\end{figure}

\textbf{Data-dependent transition model:} In order to keep the size of tracking
graphs tractable for our inference methods, we need a heuristic to select a
sparse set of links between detection candidates across frames.  We found that
simply predicting candidate's locations in future frames via optical flow gives
very good performance.  Specifically, we first compute frame-wise optical flow
using software of~\cite{Liu2009}, then for a candidate detection $x_i$ at frame
$t_i$, we compute the mean of vertical flows and the mean of horizontal flows
within the candidate box, and use them to predict candidate's location in the
next frame $t_i+1$; for $x_i$'s predicted locations in frame $t_i+2$ we use its
newly predicted location at $t_i+1$ and candidate's original box size to repeat
the process described above, and same for $t_i+n$.

\textbf{Trajectory smoothing:} During evaluation we observe that many track
fragmentation errors (FRAG) reported by the benchmark are due to the raw trajectory
oscillating away from the ground-truth due to poorly localized 
detection candidates.  Inspired by the trajectory model of
~\cite{Andriyenko:2012:DCO}, we post-process each output raw trajectory by
fitting a cubic B-spline.  This smoothing of the trajectory eliminates many
FRAGs from the raw track, making the fragmentation number more meaningful
when compared across different models.

\textbf{Baselines:} We use the publicly available code
from~\cite{Geiger2014PAMI} as a first baseline.  It relies on a three-stages
tracklet linking scheme with occlusion sensitive appearance learning; it is by
far the best tracker for cars on KITTI tracking benchmark among all published
methods. Also we consider dynamic programming (DP) and successive shortest path (SSP) with default parameters
in~\cite{PirsiavashRF_CVPR_2011} as another two baselines, denoted as DP+Flow
and SSP+Flow in our table.

\textbf{Parameter settings:} We tuned the structural parameters of the 
various baselines to give good performance.  For all baselines we only use
detections that have a positive score. For DP+Flow and SSP+Flow we also remove
all transition links that have overlap ratios lower than 0.5. For learned
tracking models (+Struct) we use detections that have scores greater than -0.5,
and transition links that have overlap ratios greater than 0.3.

\textbf{Benchmark Results:} 
We evaluate performance using a standard battery of performance measures.
The evaluation result for each object category, as well as for all categories are shown in Table~\ref{tab:tab1}.
For our learned tracking models (+Struct) we use either network simplex solver
(for SSP+Flow+Struct) or LP relaxation (for LP/DP+Flow+Struct) for
training and conduct leave-one-sequence-out cross-validation with
$C=2^{-9},2^{-8},...,2^3$. We report cross-validation result under best $C$, which is
$C=2^{-8}$ for SSP+Flow+Struct and $C=2^{-7}$ for LP/DP+Flow+Struct. Our simple
motion model helps DP+Flow outperform state-of-the-art baseline by a
significant margin.  One exception is IDSW which we attribute to the fact 
that the network-flow methods do not explicitly model target appearance.
While SSP+Flow seems to perform poorly with default parameters, it turns out
that with properly learned parameters (SSP+Flow+Struct), it produces results
that are comparable to (and often better than) DP+Flow, this indicates that
there is much more potential of SSP than suggested in previous work.  In addition,
SSP's guarantee of optimality makes it very attractive if more complicated
features and network structure are to be used in learning.  As shown in 
Table~\ref{tab:tab1}, in our evaluation over all objects our model learned with
pairwise costs (LP/DP+Flow+Struct) achieves the best MOTA, Recall, Mostly
Tracked(MT) and Mostly Lost(ML) performance while keeping other metrics
competitive.

\textbf{Approximate Inference:} To evaluate quality of the LP+rounding and DP
approximation, we run both LP+rounding and DP inference on models trained via
LP relaxation and DP respectively. We then average the running time and
minimum cost found on each sequence for LP+rounding and DP, respectively. Fig~\ref{fig:fig5}
shows the accumulative running time and cost for each algorithm. During our
experiments, LP+rounding often finds the exact relaxed global optimal, and when
it doesn't it still gives very close approximation. While greedy forward
search using DP rarely reach relaxed global optimum, it still produced good
approximate solutions that were often within 1\% of relaxed global optimum
while running significantly faster (2-7x) than LP+rounding.

\begin{table}
\begin{center}
\renewcommand{\tabcolsep}{3.5pt}
{\scriptsize
\begin{tabular}{ c | c | c | c | c | c | c | c | c |}
\bb{Car} & MOTA & MOTP & Rec & Prec & MT & ML & IDSW & FRAG \\
\hline\hline
Baseline~\cite{Geiger2014PAMI} & 57.8 & 78.8 & 58.6 & 98.8 & 14.9 & 28.4 & 22 & 225 \\
\hline\hline
SSP+Flow & 49.0 & 79.1 & 49.1 & 99.7 & 18.4 & 59.9 & 0 & 47 \\
\hline
DP+Flow & 62.2 & 79.0 & 63.4 & 98.5 & 25.2 & 24.2 & 43 & 177 \\
\hline\hline
SSP+Flow+Struct & 63.4 & 78.3 & 65.4 & 97.1 & 27.4 & 20.0 & 2 & 179 \\
\hline
LP+Flow+Struct & 64.1 & 78.1 & 67.1 & 95.7 & 30.5 & 18.7 & 3 & 208 \\
\hline
DP+Flow+Struct & 64.6 & 78.0 & 67.5 & 96.0 & 30.1 & 18.6 & 17 & 222 \\
\hline
\\
\end{tabular}}

{\scriptsize
\begin{tabular}{ c | c | c | c | c | c | c | c | c |}
\bb{Pedestrian} & MOTA & MOTP & Rec & Prec & MT & ML & IDSW & FRAG \\
\hline\hline
Baseline & 40.2 & 73.2 & 49.0 & 86.6 & 4.2 & 32.2 & 132 & 461 \\
\hline\hline
SSP+Flow & 37.9 & 73.4 & 41.8 & 92.0 & 8.4 & 57.5 & 25 & 146 \\
\hline
DP+Flow & 49.7 & 73.1 & 57.2 & 88.9 & 18.6 & 26.3 & 46 & 260 \\
\hline\hline
SSP+Flow+Struct & 51.2 & 73.2 & 57.4 & 90.5 & 19.2 & 24.6 & 16 & 230 \\
\hline
LP+Flow+Struct & 52.6 & 72.9 & 60.2 & 89.2 & 22.2 & 21.6 & 31 & 281 \\
\hline
DP+Flow+Struct & 52.4 & 73.0 & 60.0 & 89.2 & 19.8 & 22.2 & 36 & 277 \\
\hline
\\
\end{tabular}}
{\scriptsize
\begin{tabular}{ c | c | c | c | c | c | c | c | c |}
\bb{Cyclist} & MOTA & MOTP & Rec & Prec & MT & ML & IDSW & FRAG \\
\hline\hline
Baseline & 39.0 & 81.6 & 39.6 & 99.5 & 5.4 & 37.8 & 7 & 26 \\
\hline\hline
SSP+Flow & 18.7 & 85.6 & 18.7 & 100 & 5.4 & 89.2 & 0 & 1 \\
\hline
DP+Flow & 42.4 & 81.2 & 42.5 & 100 & 18.9 & 45.9 & 2 & 5 \\
\hline\hline
SSP+Flow+Struct & 47.4 & 79.7 & 59.9 & 83.0 & 35.1 & 32.4 & 5 & 10 \\
\hline
LP+Flow+Struct & 52.3 & 79.6 & 61.1 & 88.2 & 40.6 & 27.0 & 12 & 21 \\
\hline
DP+Flow+Struct & 56.3 & 79.4 & 64.2 & 89.7 & 40.5 & 27.0 & 9 & 15 \\
\hline
\end{tabular}}
\end{center}
\renewcommand{\tabcolsep}{3.5pt}
{\scriptsize
\begin{tabular}{ c | c | c | c | c | c | c | c | c |}
\bb{All Categories} & MOTA & MOTP & Rec & Prec & MT & ML & IDSW & FRAG \\
\hline\hline
Baseline & 51.7 & 77.4 & 54.8 & 95.3 & 12.1 & 29.7 & 161 & 712 \\
\hline\hline
SSP+Flow & 44.2 & \bb{77.7} & 45.5 & \bb{97.5} & 15.6 & 60.8 & 25 & \bb{194} \\
\hline
DP+Flow & 57.6 & 77.4 & 60.5 & 95.7 & 23.5 & 25.7 & 91 & 442 \\
\hline\hline
SSP+Flow+Struct & 59.0 & 77.0 & 62.8 & 94.5 & 25.9 & 21.5 & \bb{23} & 419 \\
\hline
LP+Flow+Struct & 60.2 & 76.7 & 64.8 & 93.5 & \bb{29.2} & 19.7 & 46 & 510 \\
\hline
DP+Flow+Struct & \bb{60.6} & 76.7 & \bb{65.1} & 93.8 & 28.4 & \bb{19.7} & 62 & 514 \\
\hline
\end{tabular}}
\vspace{0.05in}
\caption{Tracking result for cars, pedestrian and cyclist categories
in the KITTI tracking benchmark and aggregate performance over all categories.
The proposed method using quadratic interactions between objects and parameters
trained using structured prediction achieves state-of-the art MOTA and is 
competitive across multiple performance measures.
}
\label{tab:tab1}
\end{table}

\begin{table}
\begin{center}
{\scriptsize
\begin{tabular}{ c c || c | c | c ||}
&& \multicolumn{3}{|c|| }{Train} \\ 
&&& DP & LP \\ 
\hline\hline
\multirow{14}{*}{Test}
&\multirow{8}{*}{DP} & MOTA & 60.5 & 60.6 \\ \cline{3-5}
&& Recall & 65.2 & 65.1 \\ \cline{3-5}
&& Precision & 93.5 & 93.8 \\ \cline{3-5}
&& MT & 28.6 & 28.4 \\ \cline{3-5}
&& ML & 20.5 & 19.7 \\ \cline{3-5}
&& IDSW & 68 & 62\\ \cline{3-5}
&& FRAG & 517 & 514 \\ \cline{2-5}
&\multirow{8}{*}{LP+round} & MOTA & 60.1 & 60.2 \\ \cline{3-5}
&& Recall & 64.9 & 64.8 \\ \cline{3-5}
&& Precision & 93.3 & 93.5 \\ \cline{3-5}
&& MT & 29.3 & 29.2 \\ \cline{3-5}
&& ML & 20.3 & 19.7 \\ \cline{3-5}
&& IDSW & 56 & 46 \\ \cline{3-5}
&& FRAG & 518 & 510 \\
\hline
\hline
\end{tabular}}
\end{center}
\caption{Performance evaluation over 21 sequences using cross validation
for different combinations of inference algorithm used during training and
test time.}
\label{tab:tab2}
\end{table}

\textbf{Overgenerating versus Undergenerating:} Previous works have shown that in
general, models trained with relaxed inference are preferable than models
trained with greedy inference. To investigate this idea in our particular
problem, we also conduct leave-one-sequence-out cross-validation using
either DP or the LP relaxation as the inference method for training. The
evaluation results under different training/testing inference combinations are
shown in Table~\ref{tab:tab2}. Notice that model
trained with the LP relaxation does slightly better in most metrics, whereas DP
stands out as a good inference algorithm at test time. Moreover, though
slightly falling behind, model trained with greedy DP is very close to
the performance of that trained with LP and thus suggests the greedy algorithm
proposed here is a very competitive inference method.

\section{Summary}
\label{sec:Summary}
We augmented the well-studied network-flow tracking model with pairwise cost, and 
proposed an end-to-end framework that jointly optimizes parameters for such model.
We extensively evaluated a traditional LP relaxation-based method and a novel greedy 
dynamic programming method for inference in the augmented network, both of which 
achieves state-of-the-art performance, while our greedy DP algorithm being 2-7x 
faster than a commercial LP solver.

\section{Acknowledgements} This work was supported by NSF DBI-1053036,
IIS-1253538 and a Google Research Award.

{\small
\bibliographystyle{ieee}
\bibliography{egbib}
}

\newpage
\section{Appendix: Multi-Pass Dynamic Programming to Approximate Successive Shortest Path}
\label{sec:app1}
\begin{figure}
\begin{center}
\resizebox{0.4\textwidth}{!}{%
\begin{tikzpicture}
	\begin{pgfonlayer}{nodelayer}
		\node [style=vertex_default] (0) at (-7, 1) {};
		\node [style=vertex_default] (1) at (-7.5, 0) {};
		\node [style=vertex_default] (2) at (-8, -1) {};
		\node [style=vertex_default] (3) at (-6, 1) {};
		\node [style=vertex_default] (4) at (-6.5, 0) {};
		\node [style=vertex_default] (5) at (-7, -1) {};
		\node [style=vertex_default] (6) at (-4, 1) {};
		\node [style=vertex_default] (7) at (-4.5, 0) {};
		\node [style=vertex_default] (8) at (-3, 1) {};
		\node [style=vertex_default] (9) at (-3.5, 0) {};
		\node [style=vertex_default] (10) at (-1, 1) {};
		\node [style=vertex_default] (11) at (-1.5, 0) {};
		\node [style=vertex_default] (12) at (0, 1) {};
		\node [style=vertex_default] (13) at (-0.5, 0) {};
		\node [style=vertex_default] (14) at (-6, 3) {S};
		\node [style=vertex_default] (15) at (-2, -3) {T};
		\node [style=vertex_default] (16) at (-1, -1) {};
		\node [style=vertex_default] (17) at (-2, -1) {};
		\node [style=vertex_default] (18) at (-4, -1) {};
		\node [style=vertex_default] (19) at (-5, -1) {};
	\end{pgfonlayer}
	\begin{pgfonlayer}{edgelayer}
		\draw [style=edge_forward,draw=red] (0) to (3);
		\draw [style=edge_forward,draw=red] (1) to (4);
		\draw [style=edge_forward,draw=red] (2) to (5);
		\draw [style=edge_forward,draw=red] (6) to (8);
		\draw [style=edge_forward,draw=red] (7) to (9);
		\draw [style=edge_forward,draw=red] (10) to (12);
		\draw [style=edge_forward,draw=red] (11) to (13);
		\draw [style=edge_forward, in=90, out=180, looseness=1.25] (14) to (0);
		\draw [style=edge_forward, in=90, out=180, looseness=1.25] (14) to (1);
		\draw [style=edge_forward, in=90, out=180, looseness=1.25] (14) to (2);
		\draw [style=edge_forward, in=90, out=0, looseness=1.25] (14) to (6);
		\draw [style=edge_forward, in=90, out=0, looseness=1.25] (14) to (7);
		\draw [style=edge_forward, in=90, out=0] (14) to (10);
		\draw [style=edge_forward, in=90, out=0, looseness=1.50] (14) to (11);
		\draw [style=edge_forward, in=180, out=-90, looseness=0.75] (5) to (15);
		\draw [style=edge_forward, in=180, out=-90] (4) to (15);
		\draw [style=edge_forward, in=180, out=270, looseness=1.25] (3) to (15);
		\draw [style=edge_forward, in=180, out=-90, looseness=0.75] (9) to (15);
		\draw [style=edge_forward, in=180, out=-90, looseness=0.50] (8) to (15);
		\draw [style=edge_forward, in=0, out=-90, looseness=1.25] (13) to (15);
		\draw [style=edge_forward, in=0, out=-90, looseness=1.25] (12) to (15);
		\draw [style=edge_forward,draw=blue] (3) to (6);
		\draw [style=edge_forward,draw=blue] (4) to (7);
		\draw [style=edge_forward,draw=blue] (5) to (7);
		\draw [style=edge_forward,draw=blue] (4) to (6);
		\draw [style=edge_forward,draw=blue] (8) to (10);
		\draw [style=edge_forward,draw=blue] (9) to (11);
		\draw [style=edge_forward,draw=blue] (8) to (11);
		\draw [style=edge_forward,draw=red] (17) to (16);
		\draw [style=edge_forward,draw=red] (19) to (18);
		\draw [style=edge_forward,draw=blue] (5) to (19);
		\draw [style=edge_forward,draw=blue] (18) to (17);
		\draw [style=edge_forward, in=90, out=0, looseness=0.75] (14) to (19);
		\draw [style=edge_forward, in=180, out=-90] (18) to (15);
		\draw [style=edge_forward, in=90, out=0, looseness=1.75] (14) to (17);
		\draw [style=edge_forward, in=0, out=-90, looseness=1.25] (16) to (15);
		\draw [style=edge_forward,draw=blue] (4) to (19);
		\draw [style=edge_forward,draw=blue] (9) to (17);
	\end{pgfonlayer}
\end{tikzpicture}
}
\end{center}
\caption{Graphical representation of network flow model from~\cite{10.1109/CVPR.2008.4587584}.
A pair of nodes (connected by red edge) represent a detection, blue edges represent possible
transitions between detections and birth/death flows are modeled by black edges. Costs $c_i$ in
Objective~\ref{eqn:mincostflow} are for red edges, $c_{ij}$ are for blue edges, $c_i^t$ and $c_i^s$
are for black edges.
To simplify our description, 
we will refer a detection edge and the two nodes associated 
with it as a "node" or "detection node". The set $V$ consists of all detection nodes
in the graph, whereas the set $E$ consists of all transition edges in the graph.}
\label{fig:fig1}
\end{figure}
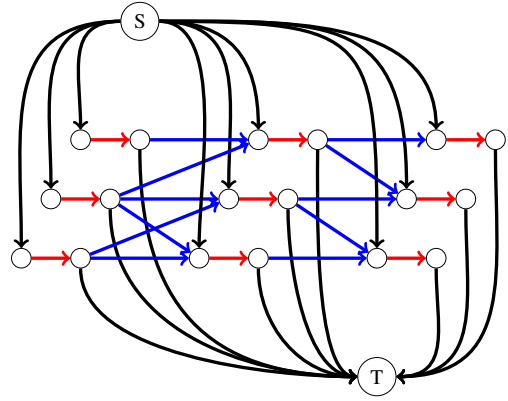

Now we describe two dynamic programming (DP) algorithms proposed
by~\cite{PirsiavashRF_CVPR_2011} which approximate successive shortest path (SSP)
algorithm. Recall the network-flow problem described in Equation~\ref{eqn:mincostflow}:
\begin{align}
\underset{\mathbf{f}}{\operatorname{min}} &\ \sum_{i} c_i^s f_i^s + \sum_{ij \in E} c_{ij} f_{ij} + \sum_{i} c_i f_i + \sum_{i} c_i^t f_i^t \nonumber
\label{eqn:mincostflow_} \\
\text{s.t.}&\quad f_i^s + \sum_j f_{ji} = f_i = f_i^t + \sum_j f_{ij} \nonumber\\
& f_i^s, f_i^t, f_i, f_{ij} \in \{0,1\} \nonumber 
\end{align}
The corresponding graphical model is shown in Fig~\ref{fig:fig1}. SSP finds the global
optimum of Objective~\ref{eqn:mincostflow} by repeating:
\begin{adjustwidth}{1em}{0pt}
1. Find the minimum cost source to sink path on residual graph $G_r(\mathbf{f})$\\
2. If the cost of the path is negative, push a flow through the path to update $\mathbf{f}$
\end{adjustwidth}
Until no negative cost path can be found. A residual graph $G_r(\mathbf{f})$ is the
same as the original graph $G$ except all edges in $\mathbf{f}$ are reversed and
their cost negated. We focus on describing the DP algorithms and refer readers
to~\cite{SSP} for detailed description of SSP algorithm.

\subsection{One-pass DP}
\label{sec:DP1}
Assume the detection nodes are sorted in time. We denote 
$cost(i)$ as the cost of the shortest path from source node to node $i$, $link(i)$ as 
$i$'s predecessor in this shortest path, and
$birth\_node(i)$ as the first detection node in this shortest path. We initialize 
$cost(i) = c_i + c_i^s$, $link(i) = \emptyset$, and $birth\_node(i) = i$ for all $i \in V$.

To find the shortest path on the initial DAG $G$, we can sweep from first frame 
to last frame, computing $cost(i)$ as:
\begin{align}
cost(i) = c_i + \text{min} (\pi, c_i^s), \; \pi = \underset{ji 
\in E}{\operatorname{min}} \; c_{ji} + cost(j)
\end{align}
And update $birth\_node(i)$, $link(i)$ accordingly.

After we sweeping through all frames, we find a node $i$ such that $cost(i) + c_i^t$ is minimum, and
reconstruct the shortest path by backtracking cached $link$ variables.
The cost of this path would be $cost(i) + c_i^t$.
After the shortest path is found, we remove all nodes and edges in this shortest path
from $G$, the resulting graph $G'$ will still be a DAG, thus we can repeat this
procedure until we cannot find any path that has a negative cost.
Even more speed up can be achieved by only recomputing $cost(i)$,
$birth\_node(i)$ and $link(i)$ for those $i$
whose birth node is the same as the birth node of the track found in previous
iteration.

It is also straightforward to integrate NMS into this algorithm:
when we pick up a shortest path, we also prune all nodes that overlap the
shortest path. In practice this "temporal NMS" can be much more aggressive than
pre-processing NMS, since the confidence of a track being composed of true
positives is much higher than single detections.

\subsection{Two-pass DP}
\label{sec:DP2}
2-pass DP works very similarly to successive shortest
path, the only difference is that instead of using Dijkstra's algorithm, we use two passes of
dynamic programming to approximate shortest path on the residual graph $G_r(\mathbf{f})$.
We denote $V_{forward}$ as the set of forward nodes in current residual graph, and 
$V_{backward}$ as the set of backward nodes in current residual graph, we describe one
iteration of 2-pass DP as below:
\begin{adjustwidth}{1em}{0pt}
1. Ignore all backward edges (including reversed detection edges) and perform one pass of 
forward DP (from first frame to last frame) on all nodes. For each node $i$, there will be 
a $path(i)$ array that stores mininum-cost source to $i$ path, with $cost(i)$ being the
total cost of this path.\\
2. Use $cost(i)$ from step 1 as initial values and perform one pass of
backward DP (from last frame to first frame) on $V_{backward}$. After this,
$cost(i)$ for $i \in V_{backward}$ would be the $cost(j) - c_{ij}$, where $j$ is
$i$'s best (backward) predecessor and $c_{ij}$ is from the original graph. Set
$cost(i) = +\infty$ for backward node $i$ that has no backward edge coming to it.\\
3. Perform one pass of forward DP on $i \in V_{forward}$. To avoid running into cyclic path,
we need to backtrack shortest paths for all $j \in N(i)$, where $N(i)$ is all neighboring
nodes that are connected to $i$ via a forward edge.
\\
4. Find node $i$ with minimum $cost(i) + c_i^t$, the (approximate) shortest path is
then $path(i)$.\\
5. Update solution $\mathbf{f}$ by setting all forward variables along
$path(i)$ to 1 and all backward variables along $path(i)$ to 0.
\end{adjustwidth}
It is straightforward to show that during the first iteration, 1-pass DP and 2-pass DP
behave identically. Also, the path found by 2-pass DP will never go into a source node
or go out of a sink node, thus in each iteration we generate exactly one more track, either by
splitting a previously found track, or by choosing a entirely new track. Therefore the algorithm
will terminate after at most $| V |$ iterations.

\section{Appendix: Incorporating Quadratic Interactions in Multi-pass DP}
\label{sec:app2}
Recall the augmented network-flow problem with quadratic cost (Eqn.~\ref{eqn:quadflow}):
\begin{align}
\begin{split}
\underset{\mathbf{f}}{\operatorname{min}}& \sum_{i} c_i^s f_i^s + \sum_{ij \in E} c_{ij} f_{ij} + \sum_{i} c_i f_i \nonumber \\
&+ \sum_{ij \in EC } q_{ij} f_i f_j + \sum_{i} c_i^t f_i^t
\end{split} \label{eqn:quadflow_} \\
\text{s.t.}&\quad f_i^s + \sum_j f_{ji} = f_i = f_i^t + \sum_j f_{ij} \nonumber\\
& f_i^s, f_i^t, f_i, f_{ij} \in \{0,1\} \nonumber 
\end{align}
Where $EC = \{ij : t_i = t_j\}$. We propose two new variants of DP algorithm
that can approximately minimize the Objective~\ref{eqn:quadflow}. They
are also divided into 1-pass DP and 2-pass DP. Since we already described
1-pass DP with pairwise interactions in the paper, we will focus on 2-pass DP with pairwise interactions here.

\subsection{Two-pass DP with quadratic interactions}
A feasible solution $\mathbf{f}$ on the network corresponds to a residual graph 
$G_r(\mathbf{f})$. We could apply the steps described in section \ref{sec:DP2} to 
find an approximate shortest path. This path may consist of both forward nodes and backward nodes, 
which correspond to uninstanced detections (but will be instanced after this iteration)
and already instanced detections (but will be uninstanced after this iteration) respectively.
We then update the (unary) cost of other nodes by adding or subtracting the pairwise cost imposed by
turning on or off selected nodes on the path. Additionally, at step 3 of 2-pass DP, one could
also consider the pairwise cost to current node imposed by previously selected nodes in the same path.
The entire procedure is described as Algorithm~\ref{alg:alg2}.
\begin{algorithm}{}
\begin{minipage}{0.45\textwidth}
\caption{Two-pass DP with pairwise Cost Update}
\label{alg:alg2}
\begin{algorithmic}[1]
\State \textbf{Input}: A Directed-Acyclic-Graph $G$ with node and edge weights $c_i, c_{ij}$
\State initialize $\mathbf{f} = \mathbf{0}$
\Repeat
  \State Find start-to-end min-cost unit flow $\mathbf{f}^*$ on $G_r(\mathbf{f})$
  \State $track\_cost = cost(\mathbf{f}^*)$
  \If {$track\_cost < 0$}
    \ForAll{$f_i \in \mathbf{f}^*$}
      \If {$f_i = 0$}
        \State $c_j = c_j + q_{ij} + q_{ji}, \forall ij, ji \in EC$
      \Else
        \State $c_j = c_j - q_{ij} - q_{ji}, \forall ij, ji \in EC$
      \EndIf
    \EndFor
    \State $\mathbf{f}^* = \neg \mathbf{f}^*$
  \EndIf
\Until{$track\_cost \ge 0$}
\State \textbf{Output}: Solution $\mathbf{f}$
\end{algorithmic}
\end{minipage}
\end{algorithm}

Notice that, to simplify our notation, we construct temporary residual graph at
the beginning of each iteration and do not negate edge weights in the original
graph. In practice, we can instead update edge costs and directions on the
original graph at the end of each iteration,
in such a case we should add pairwise costs to forward nodes or subtract pairwise costs from
backward nodes if we turn on some node, similarly we subtract pairwise costs
from forward nodes or add pairwise costs to backward nodes if we turn off some
node.

\subsection{Approximation Quality of Two-pass DP}
We found that 2-pass DP often finds lower cost than 1-pass DP but still not as good
as LP+rounding. It also runs significantly slower, even slower than LP+rounding
on long sequences. On a 1059 frame-long video with 3 categories of objects,
2-pass DP uses about 6 minutes to finish, whereas 1-pass DP finishes within 1
minute and LP+rounding finishes within 4 minutes. The leave-one-sequence-out
cross-validation result using 2-pass DP gets a MOTA of 60.4\%, which is
equivalent to that of 1-pass DP and LP relaxation.

We observe that most of the
running time for 2-pass DP is on the second forward pass, which involves backtracking
for each forward node to avoid cyclic path. It should be noted that with proper data structure
such as a hash linked-list to cache $path$ arrays, checking cyclic path can be done in
$O(1)$. Also, in the second forward pass, one could set all backward nodes
as active and propagate active labels to other forward nodes, so eventually we
might not need to look at every forward node. Overall, though showing some
incompetence in running time in our current implementation, 2-pass DP
should still be a promising inference method with better choice of
data-structures and moderate optimization.

\begin{figure}[t]
\begin{subfigure}[t]{0.25\textwidth}
\resizebox{0.95\textwidth}{!}{%
\begin{tikzpicture}
	\begin{pgfonlayer}{nodelayer}
		\node [style=vertex_default] (0) at (-7, 1) {};
		\node [style=vertex_default] (1) at (-7.5, 0) {};
		\node [style=vertex_default] (2) at (-8, -1) {};
		\node [style=vertex_default] (3) at (-6, 1) {};
		\node [style=vertex_default] (4) at (-6.5, 0) {};
		\node [style=vertex_default] (5) at (-7, -1) {};
		\node [style=vertex_default] (6) at (-4, 1) {};
		\node [style=vertex_default] (7) at (-4.5, 0) {};
		\node [style=vertex_default] (8) at (-3, 1) {};
		\node [style=vertex_default] (9) at (-3.5, 0) {};
		\node [style=vertex_default] (10) at (-1, 1) {};
		\node [style=vertex_default] (11) at (-1.5, 0) {};
		\node [style=vertex_default] (12) at (0, 1) {};
		\node [style=vertex_default] (13) at (-0.5, 0) {};
		\node [style=vertex_default] (14) at (-6, 3) {S};
		\node [style=vertex_default] (15) at (-2, -3) {T};
		\node [style=vertex_default] (16) at (-1, -1) {};
		\node [style=vertex_default] (17) at (-2, -1) {};
		\node [style=vertex_default] (18) at (-4, -1) {};
		\node [style=vertex_default] (19) at (-5, -1) {};
	\end{pgfonlayer}
	\begin{pgfonlayer}{edgelayer}
		\draw [style=edge_forward] (0) to (3);
		\draw [style=edge_forward] (1) to (4);
		\draw [style=edge_forward] (2) to (5);
		\draw [style=edge_forward] (6) to (8);
		\draw [style=edge_forward] (7) to (9);
		\draw [style=edge_forward] (10) to (12);
		\draw [style=edge_forward] (11) to (13);
		\draw [style=edge_forward, in=90, out=180, looseness=1.25] (14) to (0);
		\draw [style=edge_forward, in=90, out=180, looseness=1.25] (14) to (1);
		\draw [style=edge_forward, in=90, out=180, looseness=1.25] (14) to (2);
		\draw [style=edge_forward, in=90, out=0, looseness=1.25] (14) to (6);
		\draw [style=edge_forward, in=90, out=0, looseness=1.25] (14) to (7);
		\draw [style=edge_forward, in=90, out=0] (14) to (10);
		\draw [style=edge_forward, in=90, out=0, looseness=1.50] (14) to (11);
		\draw [style=edge_forward, in=180, out=-90, looseness=0.75] (5) to (15);
		\draw [style=edge_forward, in=180, out=-90] (4) to (15);
		\draw [style=edge_forward, in=180, out=270, looseness=1.25] (3) to (15);
		\draw [style=edge_forward, in=180, out=-90, looseness=0.75] (9) to (15);
		\draw [style=edge_forward, in=180, out=-90, looseness=0.50] (8) to (15);
		\draw [style=edge_forward, in=0, out=-90, looseness=1.25] (13) to (15);
		\draw [style=edge_forward, in=0, out=-90, looseness=1.25] (12) to (15);
		\draw [style=edge_forward] (3) to (6);
		\draw [style=edge_forward] (4) to (7);
		\draw [style=edge_forward] (5) to (7);
		\draw [style=edge_forward] (4) to (6);
		\draw [style=edge_forward] (8) to (10);
		\draw [style=edge_forward] (9) to (11);
		\draw [style=edge_forward] (8) to (11);
		\draw [style=edge_forward] (17) to (16);
		\draw [style=edge_forward] (19) to (18);
		\draw [style=edge_forward] (5) to (19);
		\draw [style=edge_forward] (18) to (17);
		\draw [style=edge_forward, in=90, out=0, looseness=0.75] (14) to (19);
		\draw [style=edge_forward, in=180, out=-90] (18) to (15);
		\draw [style=edge_forward, in=90, out=0, looseness=1.75] (14) to (17);
		\draw [style=edge_forward, in=0, out=-90, looseness=1.25] (16) to (15);
		\draw [style=edge_forward] (4) to (19);
		\draw [style=edge_forward] (9) to (17);
	\end{pgfonlayer}
\end{tikzpicture}
}
\caption{}
\end{subfigure}%
\begin{subfigure}[t]{0.25\textwidth}
\resizebox{0.95\textwidth}{!}{%
\begin{tikzpicture}
	\begin{pgfonlayer}{nodelayer}
		\node [style=vertex_default] (0) at (-7, 1) {};
		\node [style=vertex_default] (1) at (-7.5, 0) {};
		\node [style=vertex_default] (2) at (-8, -1) {};
		\node [style=vertex_default] (3) at (-6, 1) {};
		\node [style=vertex_default] (4) at (-6.5, 0) {};
		\node [style=vertex_default] (5) at (-7, -1) {};
		\node [style=vertex_default] (6) at (-4, 1) {};
		\node [style=vertex_default] (7) at (-4.5, 0) {};
		\node [style=vertex_default] (8) at (-3, 1) {};
		\node [style=vertex_default] (9) at (-3.5, 0) {};
		\node [style=vertex_default] (10) at (-1, 1) {};
		\node [style=vertex_default] (11) at (-1.5, 0) {};
		\node [style=vertex_default] (12) at (0, 1) {};
		\node [style=vertex_default] (13) at (-0.5, 0) {};
		\node [style=vertex_default] (14) at (-6, 3) {S};
		\node [style=vertex_default] (15) at (-2, -3) {T};
		\node [style=vertex_default] (16) at (-1, -1) {};
		\node [style=vertex_default] (17) at (-2, -1) {};
		\node [style=vertex_default] (18) at (-4, -1) {};
		\node [style=vertex_default] (19) at (-5, -1) {};
	\end{pgfonlayer}
	\begin{pgfonlayer}{edgelayer}
		\draw [style=edge_forward] (0) to (3);
		\draw [style=edge_forward_selected] (1) to (4);
		\draw [style=edge_forward] (2) to (5);
		\draw [style=edge_forward] (6) to (8);
		\draw [style=edge_forward_selected] (7) to (9);
		\draw [style=edge_forward] (10) to (12);
		\draw [style=edge_forward_selected] (11) to (13);
		\draw [style=edge_forward, in=90, out=180, looseness=1.25] (14) to (0);
		\draw [style=edge_forward_selected, in=90, out=180, looseness=1.25] (14) to (1);
		\draw [style=edge_forward, in=90, out=180, looseness=1.25] (14) to (2);
		\draw [style=edge_forward, in=90, out=0, looseness=1.25] (14) to (6);
		\draw [style=edge_forward, in=90, out=0, looseness=1.25] (14) to (7);
		\draw [style=edge_forward, in=90, out=0] (14) to (10);
		\draw [style=edge_forward, in=90, out=0, looseness=1.50] (14) to (11);
		\draw [style=edge_forward, in=180, out=-90, looseness=0.75] (5) to (15);
		\draw [style=edge_forward, in=180, out=-90] (4) to (15);
		\draw [style=edge_forward, in=180, out=270, looseness=1.25] (3) to (15);
		\draw [style=edge_forward, in=180, out=-90, looseness=0.75] (9) to (15);
		\draw [style=edge_forward, in=180, out=-90, looseness=0.50] (8) to (15);
		\draw [style=edge_forward_selected, in=0, out=-90, looseness=1.25] (13) to (15);
		\draw [style=edge_forward, in=0, out=-90, looseness=1.25] (12) to (15);
		\draw [style=edge_forward] (3) to (6);
		\draw [style=edge_forward_selected] (4) to (7);
		\draw [style=edge_forward] (5) to (7);
		\draw [style=edge_forward] (4) to (6);
		\draw [style=edge_forward] (8) to (10);
		\draw [style=edge_forward_selected] (9) to (11);
		\draw [style=edge_forward] (8) to (11);
		\draw [style=edge_forward] (17) to (16);
		\draw [style=edge_forward] (19) to (18);
		\draw [style=edge_forward] (5) to (19);
		\draw [style=edge_forward] (18) to (17);
		\draw [style=edge_forward, in=90, out=0, looseness=0.75] (14) to (19);
		\draw [style=edge_forward, in=180, out=-90] (18) to (15);
		\draw [style=edge_forward, in=90, out=0, looseness=1.75] (14) to (17);
		\draw [style=edge_forward, in=0, out=-90, looseness=1.25] (16) to (15);
		\draw [style=edge_forward] (4) to (19);
		\draw [style=edge_forward] (9) to (17);
	\end{pgfonlayer}
\end{tikzpicture}}
\caption{}
\end{subfigure}
\begin{subfigure}[t]{0.25\textwidth}
\resizebox{0.95\textwidth}{!}{%
\begin{tikzpicture}
	\begin{pgfonlayer}{nodelayer}
		\node [style=vertex_default, fill=Red] (0) at (-7, 1) {};
		\node [style=vertex_default] (1) at (-7.5, 0) {};
		\node [style=vertex_default, fill=Red] (2) at (-8, -1) {};
		\node [style=vertex_default, fill=Red] (3) at (-6, 1) {};
		\node [style=vertex_default] (4) at (-6.5, 0) {};
		\node [style=vertex_default, fill=Red] (5) at (-7, -1) {};
		\node [style=vertex_default, fill=Red] (6) at (-4, 1) {};
		\node [style=vertex_default] (7) at (-4.5, 0) {};
		\node [style=vertex_default, fill=Red] (8) at (-3, 1) {};
		\node [style=vertex_default] (9) at (-3.5, 0) {};
		\node [style=vertex_default, fill=Red] (10) at (-1, 1) {};
		\node [style=vertex_default] (11) at (-1.5, 0) {};
		\node [style=vertex_default, fill=Red] (12) at (0, 1) {};
		\node [style=vertex_default] (13) at (-0.5, 0) {};
		\node [style=vertex_default] (14) at (-6, 3) {S};
		\node [style=vertex_default] (15) at (-2, -3) {T};
		\node [style=vertex_default, fill=Red] (16) at (-1, -1) {};
		\node [style=vertex_default, fill=Red] (17) at (-2, -1) {};
		\node [style=vertex_default, fill=Red] (18) at (-4, -1) {};
		\node [style=vertex_default, fill=Red] (19) at (-5, -1) {};
	\end{pgfonlayer}
	\begin{pgfonlayer}{edgelayer}
		\draw [style=edge_forward] (0) to (3);
		\draw [style=edge_forward] (2) to (5);
		\draw [style=edge_forward] (6) to (8);
		\draw [style=edge_forward] (10) to (12);
		\draw [style=edge_forward, in=90, out=180, looseness=1.25] (14) to (0);
		\draw [style=edge_forward, in=90, out=180, looseness=1.25] (14) to (2);
		\draw [style=edge_forward, in=90, out=0, looseness=1.25] (14) to (6);
		\draw [style=edge_forward, in=90, out=0, looseness=1.25] (14) to (7);
		\draw [style=edge_forward, in=90, out=0] (14) to (10);
		\draw [style=edge_forward, in=90, out=0, looseness=1.50] (14) to (11);
		\draw [style=edge_forward, in=180, out=-90, looseness=0.75] (5) to (15);
		\draw [style=edge_forward, in=180, out=-90] (4) to (15);
		\draw [style=edge_forward, in=180, out=270, looseness=1.25] (3) to (15);
		\draw [style=edge_forward, in=180, out=-90, looseness=0.75] (9) to (15);
		\draw [style=edge_forward, in=180, out=-90, looseness=0.50] (8) to (15);
		\draw [style=edge_forward, in=0, out=-90, looseness=1.25] (12) to (15);
		\draw [style=edge_forward] (3) to (6);
		\draw [style=edge_forward] (5) to (7);
		\draw [style=edge_forward] (4) to (6);
		\draw [style=edge_forward] (8) to (10);
		\draw [style=edge_forward] (8) to (11);
		\draw [style=edge_forward] (17) to (16);
		\draw [style=edge_forward] (19) to (18);
		\draw [style=edge_forward] (5) to (19);
		\draw [style=edge_forward] (18) to (17);
		\draw [style=edge_forward, in=90, out=0, looseness=0.75] (14) to (19);
		\draw [style=edge_forward, in=180, out=-90] (18) to (15);
		\draw [style=edge_forward, in=90, out=0, looseness=1.75] (14) to (17);
		\draw [style=edge_forward, in=0, out=-90, looseness=1.25] (16) to (15);
		\draw [style=edge_forward] (4) to (19);
		\draw [style=edge_forward] (9) to (17);
		\draw [style=edge_backward, bend right] (14) to (1);
		\draw [style=edge_backward, bend left=45] (1) to (4);
		\draw [style=edge_backward, bend left, looseness=0.75] (4) to (7);
		\draw [style=edge_backward, bend left=45] (7) to (9);
		\draw [style=edge_backward, bend right, looseness=0.75] (9) to (11);
		\draw [style=edge_backward, bend left=45] (11) to (13);
		\draw [style=edge_backward, bend right=330, looseness=0.75] (13) to (15);
	\end{pgfonlayer}
\end{tikzpicture}}
\caption{}
\end{subfigure}%
\begin{subfigure}[t]{0.25\textwidth}
\resizebox{0.95\textwidth}{!}{%
\begin{tikzpicture}
	\begin{pgfonlayer}{nodelayer}
		\node [style=vertex_default] (0) at (-7, 1) {};
		\node [style=vertex_default] (1) at (-7.5, 0) {};
		\node [style=vertex_default] (2) at (-8, -1) {};
		\node [style=vertex_default] (3) at (-6, 1) {};
		\node [style=vertex_default] (4) at (-6.5, 0) {};
		\node [style=vertex_default] (5) at (-7, -1) {};
		\node [style=vertex_default] (6) at (-4, 1) {};
		\node [style=vertex_default] (7) at (-4.5, 0) {};
		\node [style=vertex_default] (8) at (-3, 1) {};
		\node [style=vertex_default] (9) at (-3.5, 0) {};
		\node [style=vertex_default] (10) at (-1, 1) {};
		\node [style=vertex_default] (11) at (-1.5, 0) {};
		\node [style=vertex_default] (12) at (0, 1) {};
		\node [style=vertex_default] (13) at (-0.5, 0) {};
		\node [style=vertex_default] (14) at (-6, 3) {S};
		\node [style=vertex_default] (15) at (-2, -3) {T};
		\node [style=vertex_default] (16) at (-1, -1) {};
		\node [style=vertex_default] (17) at (-2, -1) {};
		\node [style=vertex_default] (18) at (-4, -1) {};
		\node [style=vertex_default] (19) at (-5, -1) {};
	\end{pgfonlayer}
	\begin{pgfonlayer}{edgelayer}
		\draw [style=edge_forward] (0) to (3);
		\draw [style=edge_forward] (2) to (5);
		\draw [style=edge_forward_selected] (6) to (8);
		\draw [style=edge_forward] (10) to (12);
		\draw [style=edge_forward, in=90, out=180, looseness=1.25] (14) to (0);
		\draw [style=edge_forward, in=90, out=180, looseness=1.25] (14) to (2);
		\draw [style=edge_forward_selected, in=90, out=0, looseness=1.25] (14) to (6);
		\draw [style=edge_forward, in=90, out=0, looseness=1.25] (14) to (7);
		\draw [style=edge_forward, in=90, out=0] (14) to (10);
		\draw [style=edge_forward, in=90, out=0, looseness=1.50] (14) to (11);
		\draw [style=edge_forward, in=180, out=-90, looseness=0.75] (5) to (15);
		\draw [style=edge_forward_selected, in=180, out=-90] (4) to (15);
		\draw [style=edge_forward, in=180, out=270, looseness=1.25] (3) to (15);
		\draw [style=edge_forward, in=180, out=-90, looseness=0.75] (9) to (15);
		\draw [style=edge_forward, in=180, out=-90, looseness=0.50] (8) to (15);
		\draw [style=edge_forward, in=0, out=-90, looseness=1.25] (12) to (15);
		\draw [style=edge_forward] (3) to (6);
		\draw [style=edge_forward] (5) to (7);
		\draw [style=edge_forward] (4) to (6);
		\draw [style=edge_forward] (8) to (10);
		\draw [style=edge_forward_selected] (8) to (11);
		\draw [style=edge_forward] (17) to (16);
		\draw [style=edge_forward] (19) to (18);
		\draw [style=edge_forward] (5) to (19);
		\draw [style=edge_forward] (18) to (17);
		\draw [style=edge_forward, in=90, out=0, looseness=0.75] (14) to (19);
		\draw [style=edge_forward, in=180, out=-90] (18) to (15);
		\draw [style=edge_forward, in=90, out=0, looseness=1.75] (14) to (17);
		\draw [style=edge_forward, in=0, out=-90, looseness=1.25] (16) to (15);
		\draw [style=edge_forward] (4) to (19);
		\draw [style=edge_forward] (9) to (17);
		\draw [style=edge_backward, bend right] (14) to (1);
		\draw [style=edge_backward, bend left=45] (1) to (4);
		\draw [style=edge_backward_selected, bend left, looseness=0.75] (4) to (7);
		\draw [style=edge_backward_selected, bend left=45] (7) to (9);
		\draw [style=edge_backward_selected, bend right, looseness=0.75] (9) to (11);
		\draw [style=edge_backward, bend left=45] (11) to (13);
		\draw [style=edge_backward, bend right=330, looseness=0.75] (13) to (15);
	\end{pgfonlayer}
\end{tikzpicture}}
\caption{}
\end{subfigure}
\begin{center}
\begin{subfigure}[t]{0.25\textwidth}
\resizebox{0.95\textwidth}{!}{%
\begin{tikzpicture}
	\begin{pgfonlayer}{nodelayer}
		\node [style=vertex_default] (0) at (-7, 1) {};
		\node [style=vertex_default] (1) at (-7.5, 0) {};
		\node [style=vertex_default] (2) at (-8, -1) {};
		\node [style=vertex_default] (3) at (-6, 1) {};
		\node [style=vertex_default] (4) at (-6.5, 0) {};
		\node [style=vertex_default] (5) at (-7, -1) {};
		\node [style=vertex_default, fill=Blue] (6) at (-4, 1) {};
		\node [style=vertex_default, fill=Red] (7) at (-4.5, 0) {};
		\node [style=vertex_default, fill=Blue] (8) at (-3, 1) {};
		\node [style=vertex_default, fill=Red] (9) at (-3.5, 0) {};
		\node [style=vertex_default] (10) at (-1, 1) {};
		\node [style=vertex_default] (11) at (-1.5, 0) {};
		\node [style=vertex_default] (12) at (0, 1) {};
		\node [style=vertex_default] (13) at (-0.5, 0) {};
		\node [style=vertex_default] (14) at (-6, 3) {S};
		\node [style=vertex_default] (15) at (-2, -3) {T};
		\node [style=vertex_default] (16) at (-1, -1) {};
		\node [style=vertex_default] (17) at (-2, -1) {};
		\node [style=vertex_default, fill=Red] (18) at (-4, -1) {};
		\node [style=vertex_default, fill=Blue] (19) at (-5, -1) {};
	\end{pgfonlayer}
	\begin{pgfonlayer}{edgelayer}
		\draw [style=edge_forward] (0) to (3);
		\draw [style=edge_forward] (2) to (5);
		\draw [style=edge_backward, bend left=45] (6) to (8);
		\draw [style=edge_forward] (10) to (12);
		\draw [style=edge_forward, in=90, out=180, looseness=1.25] (14) to (0);
		\draw [style=edge_forward, in=90, out=180, looseness=1.25] (14) to (2);
		\draw [style=edge_backward, in=165, out=270] (14) to (6);
		\draw [style=edge_forward, in=90, out=0, looseness=1.25] (14) to (7);
		\draw [style=edge_forward, in=90, out=0] (14) to (10);
		\draw [style=edge_forward, in=90, out=0, looseness=1.50] (14) to (11);
		\draw [style=edge_forward, in=180, out=-90, looseness=0.75] (5) to (15);
		\draw [style=edge_backward, in=180, out=-60] (4) to (15);
		\draw [style=edge_forward, in=180, out=270, looseness=1.25] (3) to (15);
		\draw [style=edge_forward, in=180, out=-90, looseness=0.75] (9) to (15);
		\draw [style=edge_forward, in=180, out=-90, looseness=0.50] (8) to (15);
		\draw [style=edge_forward, in=0, out=-90, looseness=1.25] (12) to (15);
		\draw [style=edge_forward] (3) to (6);
		\draw [style=edge_forward] (5) to (7);
		\draw [style=edge_forward] (4) to (6);
		\draw [style=edge_forward] (8) to (10);
		\draw [style=edge_backward, in=191, out=-79, looseness=0.75] (8) to (11);
		\draw [style=edge_forward] (17) to (16);
		\draw [style=edge_forward] (19) to (18);
		\draw [style=edge_forward] (5) to (19);
		\draw [style=edge_forward] (18) to (17);
		\draw [style=edge_forward, in=90, out=0, looseness=0.75] (14) to (19);
		\draw [style=edge_forward, in=180, out=-90] (18) to (15);
		\draw [style=edge_forward, in=90, out=0, looseness=1.75] (14) to (17);
		\draw [style=edge_forward, in=0, out=-90, looseness=1.25] (16) to (15);
		\draw [style=edge_forward] (4) to (19);
		\draw [style=edge_forward] (9) to (17);
		\draw [style=edge_backward, bend right] (14) to (1);
		\draw [style=edge_backward, bend left=45] (1) to (4);
		\draw [style=edge_forward] (4) to (7);
		\draw [style=edge_forward] (7) to (9);
		\draw [style=edge_forward] (9) to (11);
		\draw [style=edge_backward, bend left=45] (11) to (13);
		\draw [style=edge_backward, bend right=330, looseness=0.75] (13) to (15);
	\end{pgfonlayer}
\end{tikzpicture}}
\caption{}
\end{subfigure}
\end{center}
\caption{An illustration for 2-pass DP with quadratic interactions. (a) the
initial DAG graph, a pair of nodes indicate a candidate detection; (b) first
iteration of the algorithm, red edges indicates the shortest path found in this
iteration; (c) we reverse all the edges on the shortest path (green edges), and
add the pairwise cost imposed by this path to other candidates within the time
window (red pairs); (d) second iteration of algorithm, red edges and blue edges indicates the
new shortest path, notice that it takes three of reversed edges (blue edges);
(e) we again reverse all the edges in the shortest path, now green edges
indicate the two tracks found in this 2 iterations; we also update pairwise
cost: blue node pair indices we subtract the pairwise cost imposed by "turning off"
an candidate, red pair still indicates adding in pairwise cost of newly
instanced candidates,and the blue-red pair indicates we first add the pairwise
cost by newly instanced candidates, then subtract the pairwise cost by newly
uninstanced candidates. Additions and subtractions are done to the non-negated
edge costs and then negated if necessary.}
\end{figure}
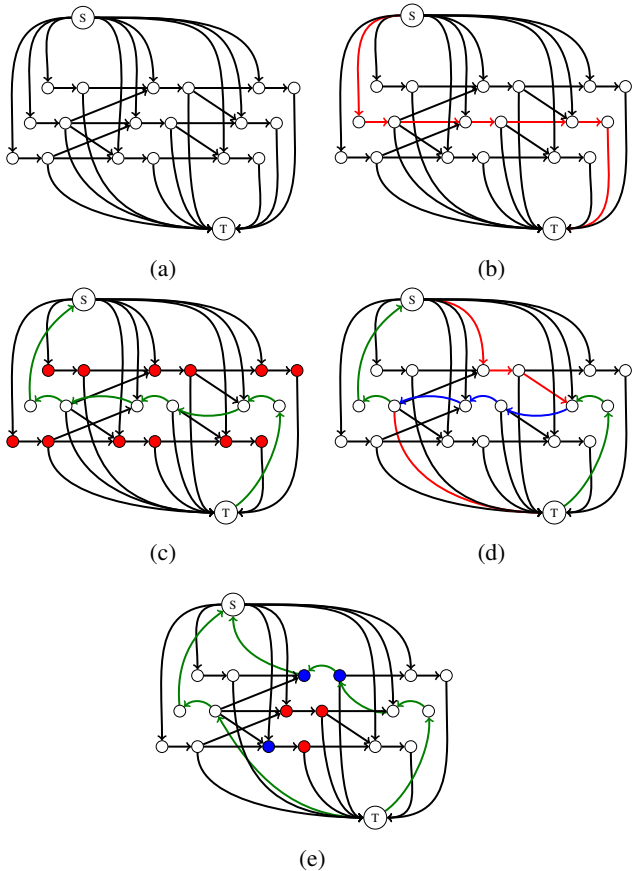

\end{document}